\documentclass[letterpaper]{ieeeconf}

\usepackage{graphicx}
\usepackage{lineno}
\usepackage{amsmath}
\usepackage{hyperref}

\usepackage{subcaption}
\usepackage{graphicx}
\usepackage{afterpage}
\usepackage[left=0.98in, right=0.98in, top=1.0in, bottom=1.0in]{geometry}

\newcommand\beq{\begin{equation}}
\newcommand\eeq{\end{equation}}

\newcommand\bmat{\begin{bmatrix}}
\newcommand\emat{\end{bmatrix}}

\begin{document}
\setpagewiselinenumbers        %  Line numbers for edits to drafts.
\modulolinenumbers[1]          %  number every N lines

% \linenumbers                   %  start numbering lines here

\title{Optimization of Trajectories for Machine Learning Training in Robot Accuracy Modeling}

\author{
        Blake Hannaford\\
        Department of Electrical \& Computer Engineering \\
        The University of Washington\\
       \today
 }

\maketitle

%%%%** Section 1 
\section{Introduction}
Among the exciting applications of Machine Learning (ML) in robot control
is the possibility of increasing the accuracy of flexible and friction-prone
control of robotic joints in elongated mechanisms.   Surgical instruments are an
extreme example of robotic end effectors which require a very high length to thickness
aspect ratio of the links (especially more distal links).   This premium is especially
marked for endoscopic surgical instruments as in the daVinci surgical robotic
system from Intuitive Surgical Inc.

With long thin mechanisms, in which it is impractical to mount actuators directly
on the joints (also a property of robotic surgical  instruments), transmission elements
undergo large strains and stresses.   Elements such as cables, torsion rods, and associated
pullies and tiny bearings inevitably introduce properties such as elastic deformation
and backlash (position error), and friction. Friction can be though of as
velocity dependent force.   At the low speeds typical of robotic surgical
applications, friction forces are large and highly non-linear.

Machine learning has been recently applied to model these effects. To train such models
a robot is moved through various training points, and ground truth position and force
data are acquired through high accuracy sensors of the robot's end-point
\cite{haghighipanah2017utilizing, miyasaka2016hysteresis,  peng2020real, hwang2020efficiently}.

It is well known that modern ML methods require large sets of training data and that
acquisition of useful training data sets from real physical robots is expensive.
A workspace target is defined (seldom as big as the full robot's work volume) and
it can be filled with random points or a regular grid of points.   Training data
must be collected with the robot posed in each of these points.  Since friction
varies with velocity, to fully characterize the mechanism we must also select velocities
and visit each of the positions at each of the velocities.
This can easily grow to a very large number of points, but it takes time and energy
for the robot to move between these points and velocities.

Given a set of points in space, and a set of velocities in the same space, we
define the {\it phase space} as the combined space of positions and velocities.

We can summarize the problem of this paper as:
Given a set of points in a phase space, what is the lowest cost trajectory through
the phase space which visits each point exactly once?   For a trajectory between
two points in phase space, we define cost in this
paper as either the duration of the trajectory or the energy of the trajectory.

This is of course a variation of the Traveling Salesman Problem.

%%%%** Section 1.1
\subsection{Goals}

Specifically, the goals are
\begin{itemize}
  \item Study the difficulty of the TSP in this application by applying it
  to $X,Y,Z$ motion in 3D space ($6D$ phase space).
  \item Assess the effectiveness of an extremely simple Nearest Neighbor ('NN')
  search
  algorithm compared to random sampling of the search space.
\end{itemize}

%%%%** Section 1.2
\subsection{Approach}

In this work we will study a practical
`Nearest Neighbor'\cite{cirasella2001asymmetric} heuristic search for this special case
of the TSP we will designate as ``NN".  For simplicity (and because it seems to obtain good results)
we study the simple  heuristic algorithm which when building a path from a starting
point, chooses a random branch from among a set of branches found to have
approximately  the lowest cost value (within 2\%) among all unvisited nodes.
There are many improvements on this heuristic algorithm
(reviewed in \cite{cirasella2001asymmetric,johnson2007experimental,punnen2007traveling})
and we assume that some of them could further improve our results.

%%%%** Section 1.3
\subsection{Literature Review}

Recent work has applied ML to the problem of increasing the accuracy of
controlling manipulators with adverse mechanical properties such as flexibility
and friction in transmission components\cite{haghighipanah2017utilizing, miyasaka2016hysteresis, peng2020real, hwang2020efficiently}.
All of this work relies on training data collected from programmed robot
motion - inherently a slow process.  Since mechanical properties change with
wear and usage, repeated training during the robot's lifetime is desirable.

Assuming that there is a set of points in space, and discrete velocities, which
could sample the robot's workspace adequately for ML correction, we need to
visit those points as fast as possible.
The TSP is one approach to optimizing the visiting of training points.

The TSP is a very well studied problem with many applications in, for example,
logistics, machine scheduling, PC board drilling, and X-ray crystallography\cite{punnen2007traveling}.
Notably, \cite{bland1989large} applied
TSP to motion control of a 3-axis mechanical X-ray source in order to collect the required
images as efficiently as possible.  However this application requires 0 velocity
at each point and is symmetric.

TSP computational approaches
have been recently reviewed in \cite{rego2002traveling, punnen2007traveling}.
Most often, the TSP is studied
in 2D Euclidian space with symmetric costs (such as physical distance)
between nodes.

Few papers seem to have studied a TSP in which points share components between
position and velocity (phase space).
An exception is \cite{adler2023traveling} which derives theoretical lower bounds
for a class of TSPs that includes double integrator systems which are somewhat
representative of the system used here.
Because of the linkage between position and
velocity, trajectories in phase space between two points must be asymmetric.
% One exception is \cite{Physics Phase Space TSP}
% which ***************************.

%%%%** Section 2 
\section{Methods}

We verified that our problem is asymmetric by computing that randomized
phase space trajectories differed in cost depending
on the direction of travel between two points.
\cite{johnson2007experimental} reviews the added difficulties imposed by
this complexity.
Using the energy cost but not the time cost (see below) the problem is
non-Euclidean because with the energy cost, we
found many point-triplets in phase space for which the triangle inequality
did not hold\cite{junger1995traveling}.

Trajectories between points were synthesized by fitting a 3rd order polynomial to
the starting and ending position and velocity (Section \ref{SecNotationBasics}).
The speed of each trajectory was scaled such that the maximum acceleration was $A_{max}=2$.

%%%%** Section 2.1
\subsection{Spaces}

Computational complexity of the TSP grows extremely quickly.  We consider
grids of $N$ points on each axis normalized to $[-1,-1]$.   The simplest case is a 2D space with
$N$ positions along a line, and $N$ velocities at each point.
A more realistic case is a 6D space consisting of $X,Y,Z$ position axes and
$\dot{X},\dot{Y},\dot{Z}$ velocity axes.  We will focus our work on these
2D and 6D cases.

We then must visit $N^m$ points where $m$ is even (in our case 2 or 6).
There are $N^m$ possible starting points and then $N^m -1 $ possibilities for
the second point, etc.  Thus there are  {\it factorial}($N^m$) possible
paths. For 2D, $m=2, N=3$, we have 9 points and 362880 possible paths.  This allows
for exhaustive search resulting in a global optimum result.  For $N=4$ this
balloons to 20922789888000 ($2.092\times10^{13}$).
Using the computation time that our generic PC took to search 362880 paths,
the predicted time for the 2D, $N=4$ exhaustive search was 220 years(!).
For $N=3, m=6$, however
the number of paths is $1.4394420\times10^{1772}$ \footnote{The number of atoms in
the universe is estimated to be $10^{82}$ (\url{https://www.livescience.com/how-many-atoms-in-universe.html}, accessed 23-Aug-23.)}.
So, except for $N=3,m=2$ exhaustive search is not feasible.
Although our code was in unoptimized Python 3, allowing for a 100x speedup for C-based
code still predicts daunting runtimes.

Initially we studied rectangular grids of points, equally spaced in position
and velocity.   To determine if there were special features of optimal trajectories
on rectangular grids, we did additional computations on random grids. These
were generated off line and stored in a file so that comparisons between different
search methods could be made on the same random grid.

We will then study 4 spaces, {\it
2D rectangular,
2D random,
6D rectangular,
6D random}.
In some cases, computations on the random grid were repeated on different
random grids to confirm observed effects were not due to a specific random configuration but
results were very similar.

%%%%** Section 2.2
\subsection{Code and computing}

Trajectory generation, cost evaluation, nearest-neighbor (NN) heuristic searching,
exhaustive searching, and random path sampling were coded in Python3 with
(among others) the {\tt numpy} and {\tt itertools} libraries.
Full source code is available on Github\footnote{\url{https://github.com/blake5634/CalTrajOpt}}.

\paragraph{Heuristic Search, Starting point dependence}\label{prob:startingpt}
Heuristic NN searches produce a result which can depend on starting point.
In our application, a robot must be initialized and then incur a small cost
(compared to the overall trajectory cost) to move to the starting point
selected for an optimal path.  However we will consider all starting points
equally.

For the 2D, 4x4 grid there are 16 starting points and for the
6D, 4x4 grid there are $4^6=4096$ starting points.

% ???? For the 4x4 grid there are 729 possible start points (16!).

As above, if during the NN search,
branch costs were within 2\% of the lowest cost branch, they
were considered equal and one was chosen randomly each time.
The largest tie (set of paths within 2\%  of optimium time cost)
encountered during the 4x4 rectangular NN time-cost
search was 56 branches.

%%%%** Figure 1 
\begin{figure}
\centering
\begin{subfigure}{0.35\textwidth}
    \includegraphics[width=\textwidth]{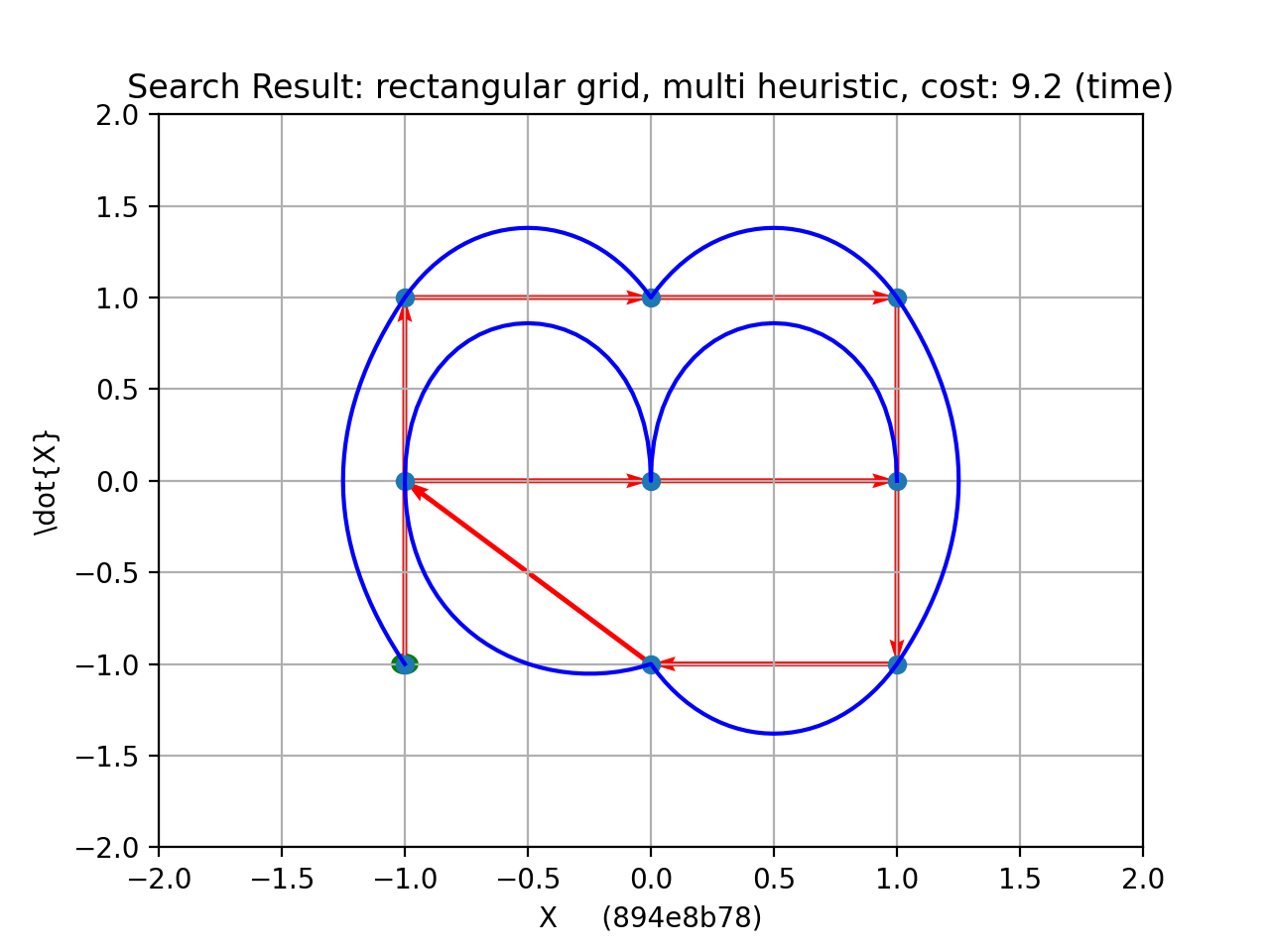}
    \caption{Suboptimal: Time Cost}
    \label{fig1:first}
\end{subfigure}
\hfill
\begin{subfigure}{0.35\textwidth}
    \includegraphics[width=\textwidth]{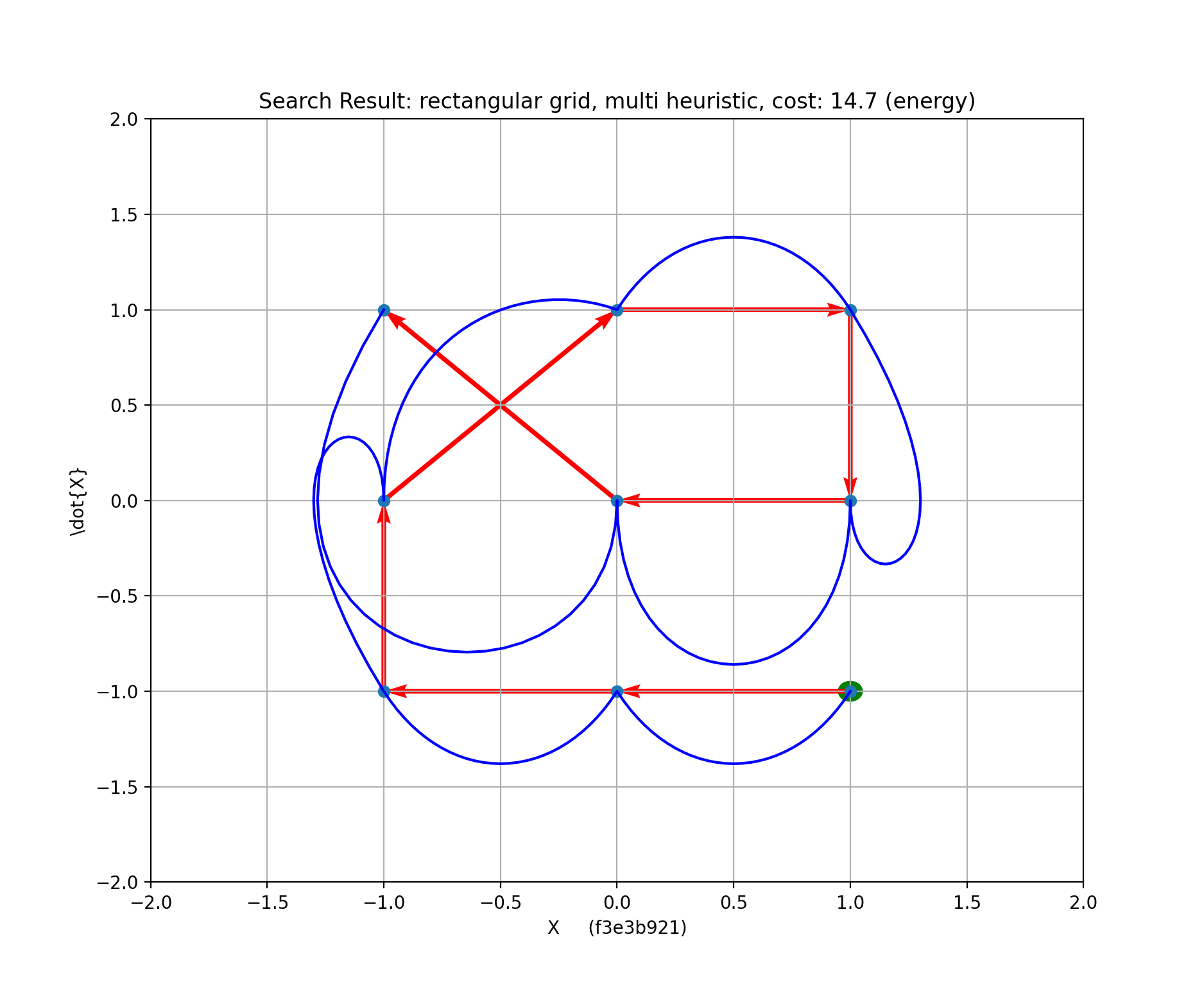}
    \caption{Suboptimal: Energy Cost}
    \label{fig1:second}
\end{subfigure}
\hfill
\begin{subfigure}{0.35\textwidth}
    \includegraphics[width=\textwidth]{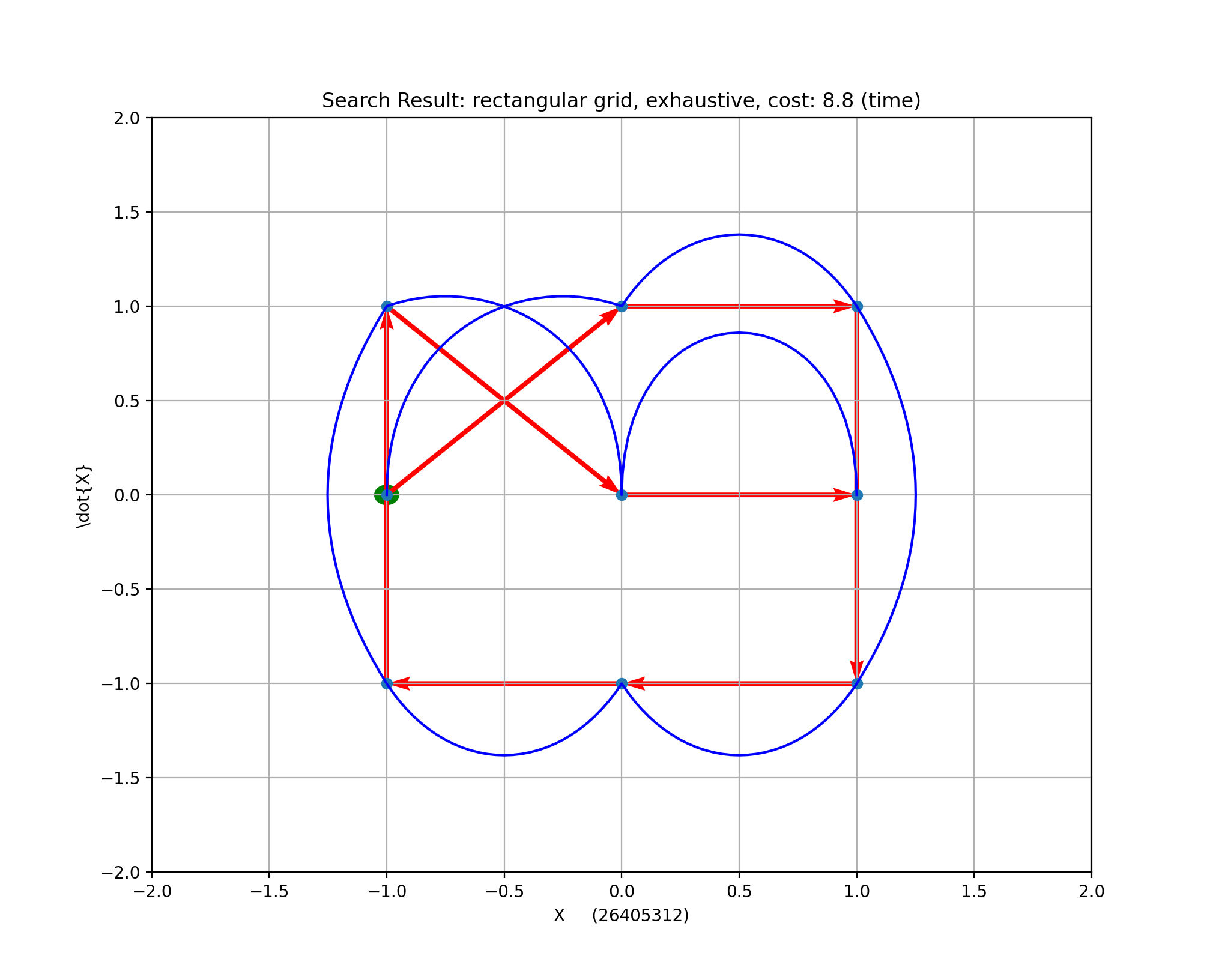}
    \caption{Globally optimal: Time Cost}
    \label{fig1:third}
\end{subfigure}
\begin{subfigure}{0.35\textwidth}
    \includegraphics[width=\textwidth]{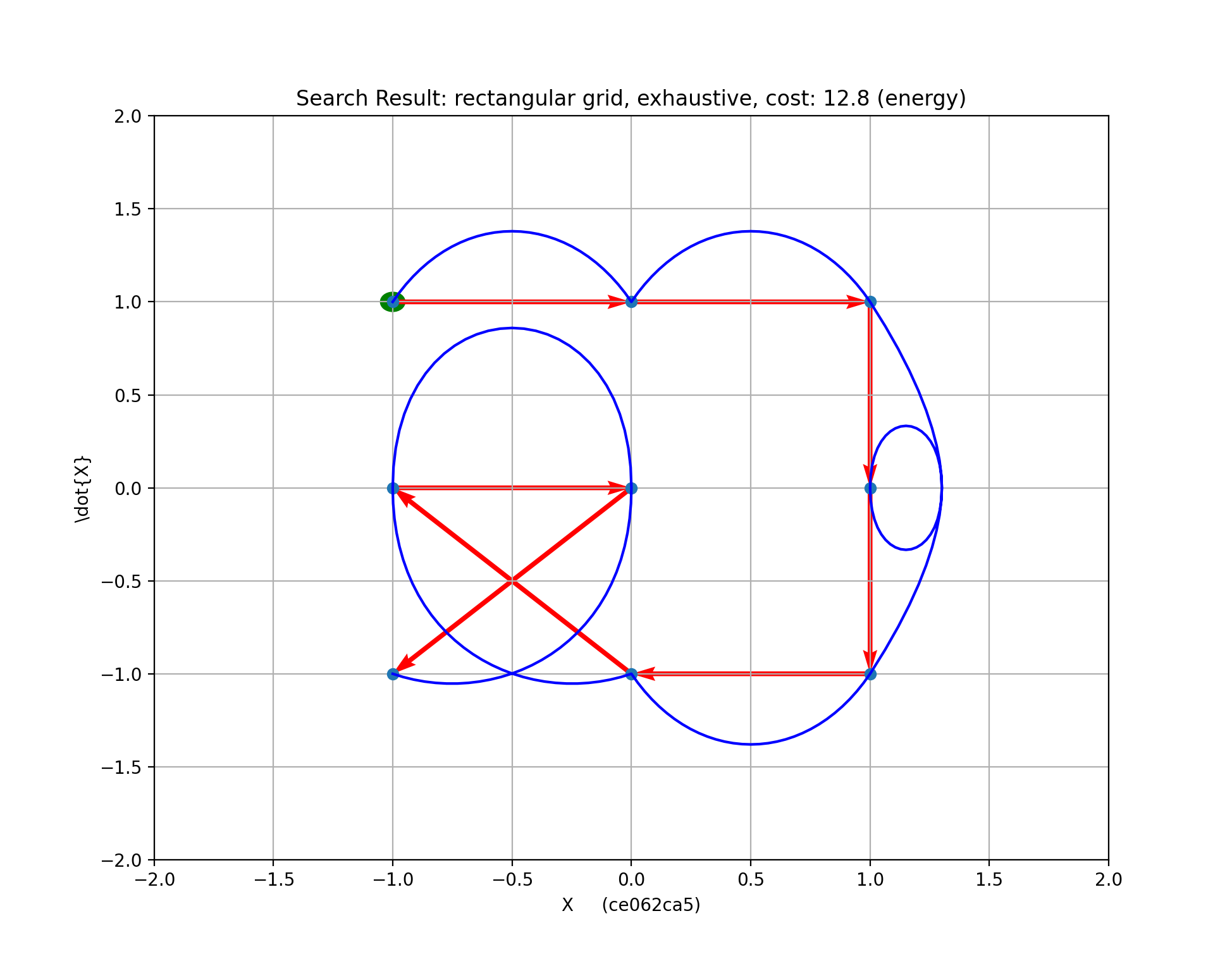}
    \caption{Globally optimal: Energy Cost}
    \label{fig1:fourth}
\end{subfigure}
%   A)\includegraphics[width=0.8\columnwidth]{redoFig2MH4ptsSubOptTime_894e8b78.png}
%   B)\includegraphics[width=0.8\columnwidth]{redoFig3MH4ptsSubOptEnergy_f3e3b921.png}
%   C)\includegraphics[width=0.8\columnwidth]{redoF2rectGr3x3ExhaustTime_26405312.png} % 8/17/23
%   D)\includegraphics[width=0.8\columnwidth]{redoF2RectGridExhaustEnergy_ce062ca5.png}
\caption{ 2D, 3x3 searches, rectangular grid.  a,b: Suboptimal paths, found by heuristic search
  (using only 4 starting points). Starting point is enlarged green circle.
  Red arrows show sequence of points, blue
  curves are the solved trajectories (Section \ref{TrajectorySolved},
  Fig. \ref{basicTraj}).  c,d: Globally optimal path on 3x3 grid found by exhaustive searches.
  Hash codes identify pertinent data files.
  }
  \label{3x3MultiHeuristicRectGrid}  \label{3x3ExhaustivRectGrid}
\end{figure}

%%%%** Section 3 
\section{Results}

%%%%** Section 3.1
\subsection{2D, $N=3$}
%%%%** Section 3.1.1 
\subsubsection{Rectangular Grid}\label{3x3RectSection}

Our first result is a very quick search using only 4 starting points of the
9 possible in the
2D, $N=3$ space and the NN search.   This was expected to get suboptimal
trajectories (Figure \ref{3x3MultiHeuristicRectGrid} top row).  Because of the sign
convention in phase space, individual point-to-point trajectories tend to
go in clockwise loops.

With an exhaustive search of the 362,880 possible paths, we found the
globally optimal trajectories minimizing time and energy costs (Figure
\ref{3x3ExhaustivRectGrid} bottom row).

\begin{figure}\centering
  \includegraphics[width=\columnwidth]{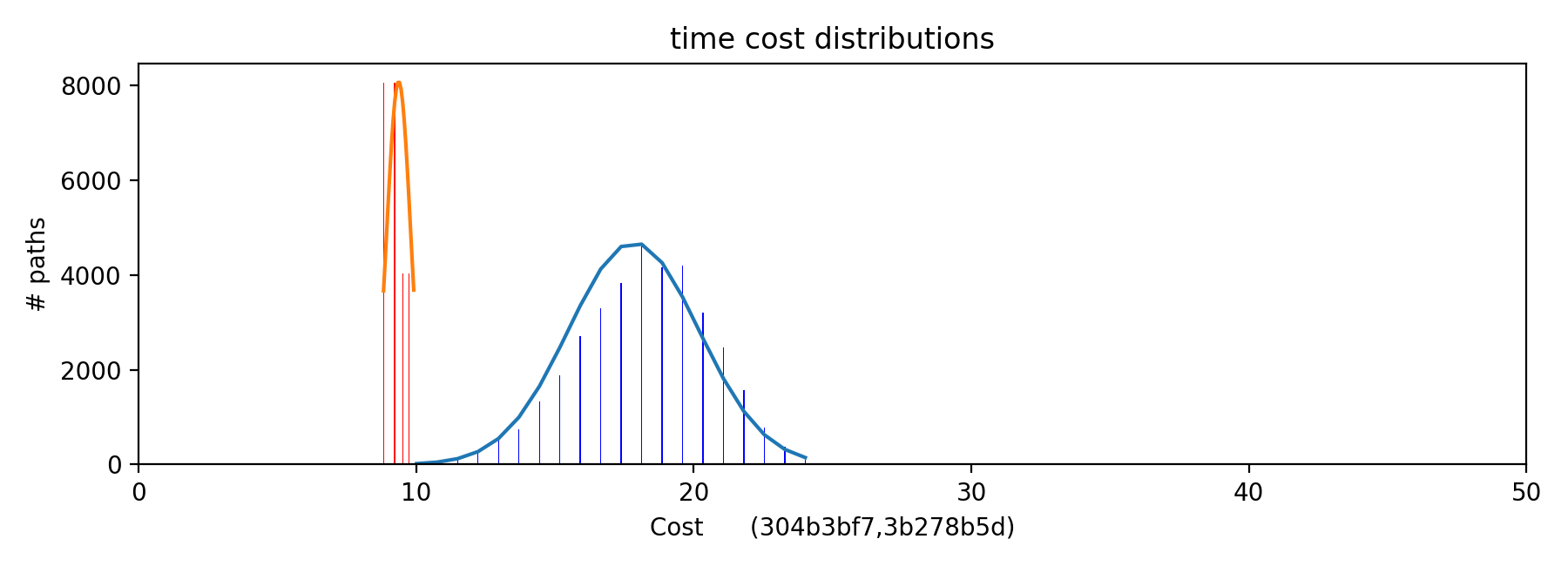}
  \includegraphics[width=\columnwidth]{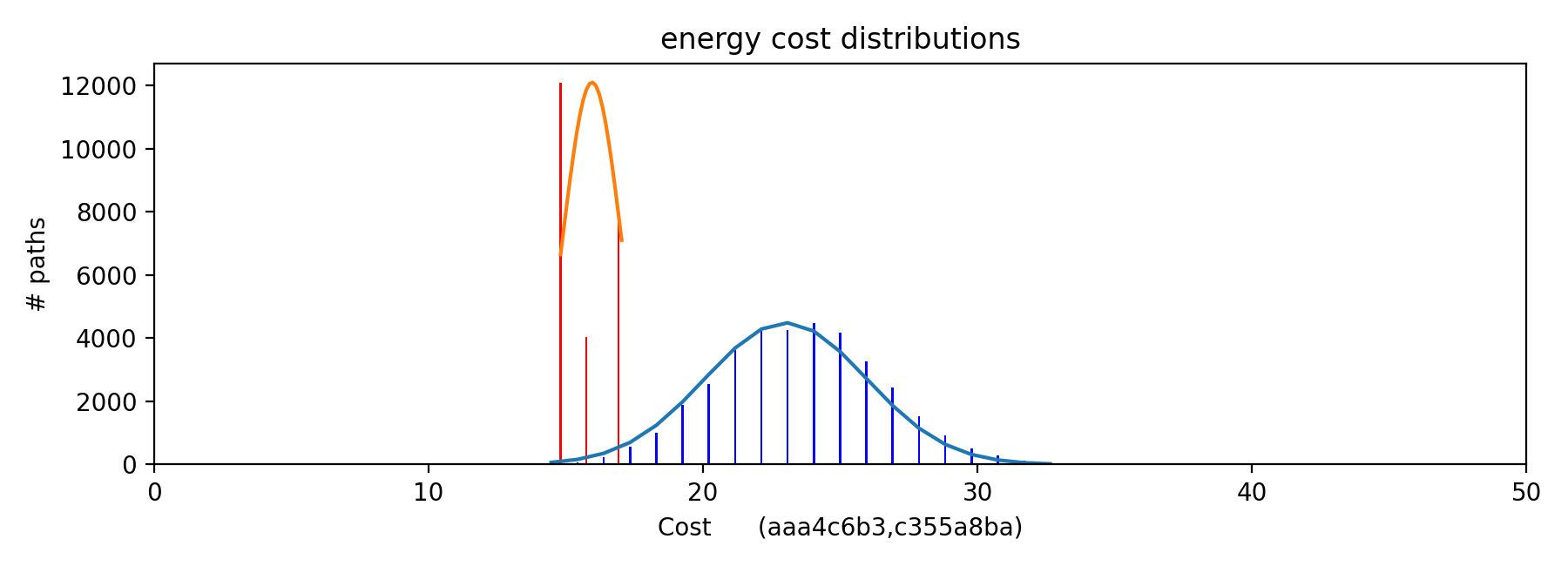}
  \caption{Comparing distributions of 10\%
  heuristic (nearest-neighbor) paths (red) with 10\% random search (blue, all 9
  starting points) through the 2-D   grid ($N=3$) by total time (Top)
  and total path energy use (Bottom).}\label{3x3x1CostDiff}
\end{figure}

Next we ran 36,288 (10\% of the total number of paths) iterations
of the NN search.    We over-plotted the distribution of costs from the multiple
NN searches for comparison with the same number of randomly selected paths
(Figure \ref{3x3x1CostDiff}).  The distribution of the 10\% random sample  was
within 1\% of the distribution of all 362,880 paths (computed but not shown).

%%%%** Figure 2 

%%%%** Section 3.1.2 
\subsubsection{Random Grid}

\begin{figure}\centering
  \includegraphics[width=\columnwidth]{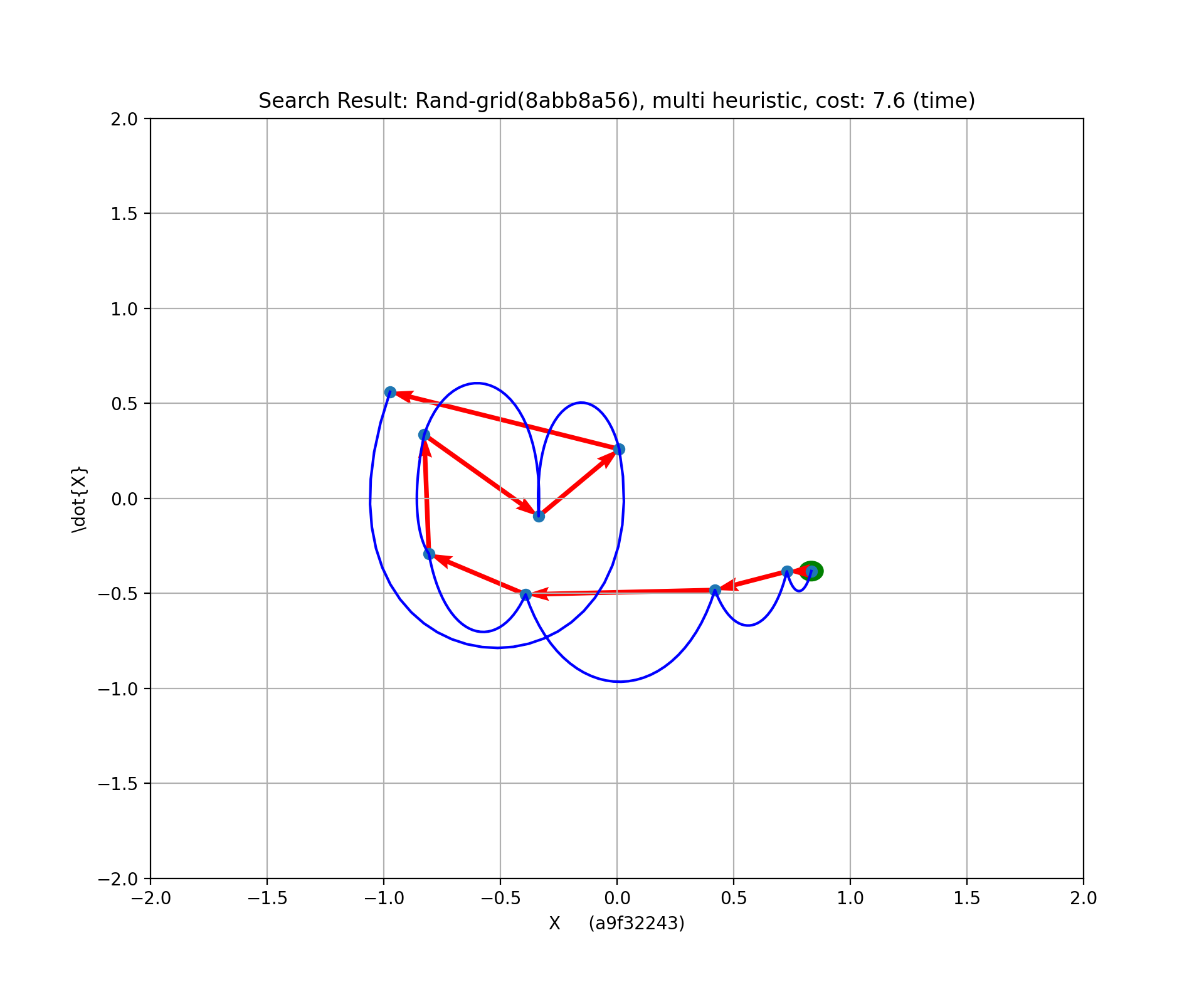}
  \includegraphics[width=\columnwidth]{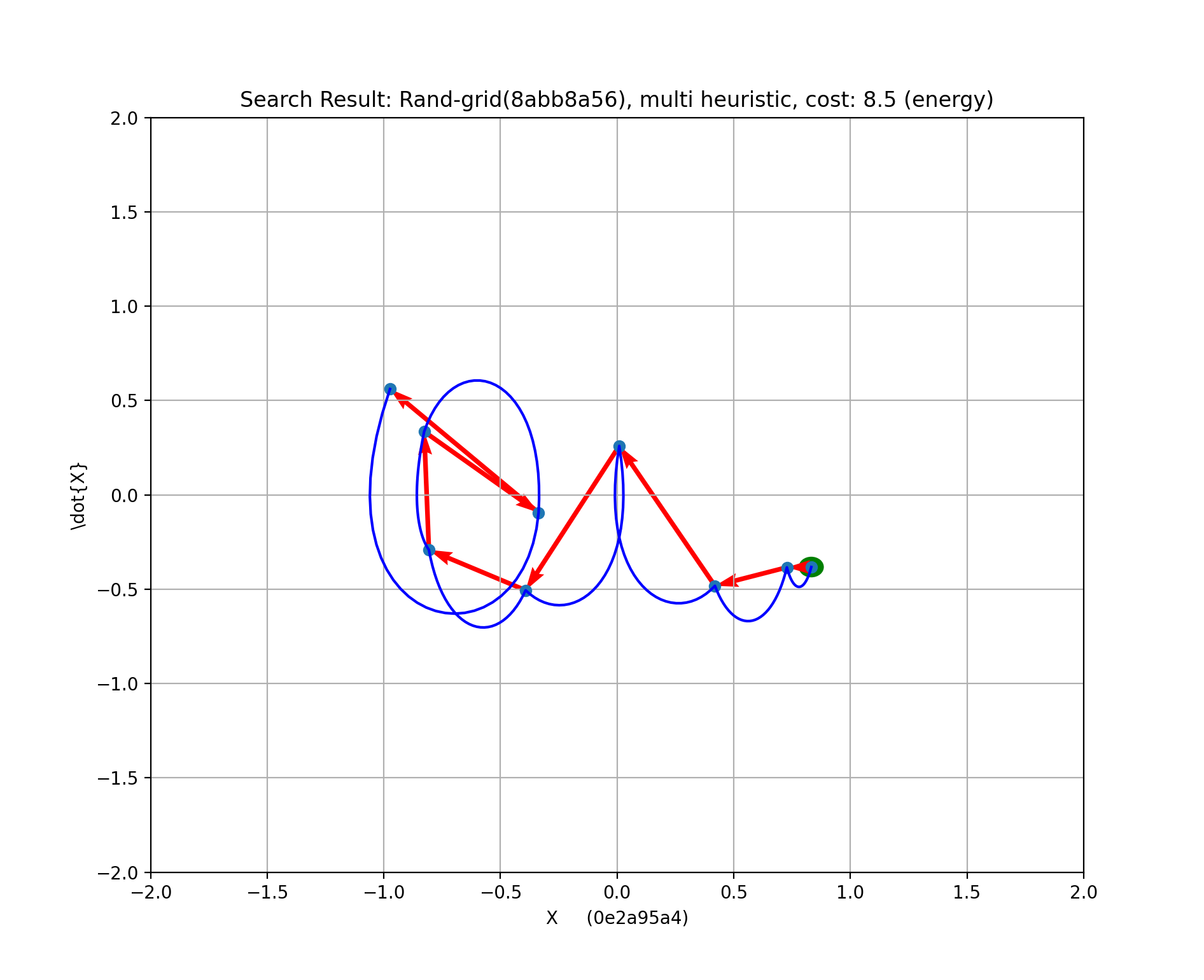}
  \caption{4 NN searches at each of   9 starting points
  in a   9-point random grid  produced
  these paths.  Path time cost (Top)
  and energy cost (Bottom). Grid points used (8abb9a56).
  }\label{3x3RandGridNN}
\end{figure}

We generated a grid of 9 random points (uniform $[-1,1]$) and ran
both NN and sampled searches.  Suboptimal paths on the random grid
are shown  in Figure \ref{3x3RandGridNN}.
Best time and energy costs for the NN searches were 7.6 and 8.5 respectively.
The corresponding global
optimal trajectories are given in Figure \ref{3x3RandGridNN}.
The trajectories appear different and their time and energy costs were lower:
6.1 and 7.0 respectively.

%%%%** Figure 3 

%%%%** Figure 4 
 \begin{figure}\centering
  \includegraphics[height=2.4in]{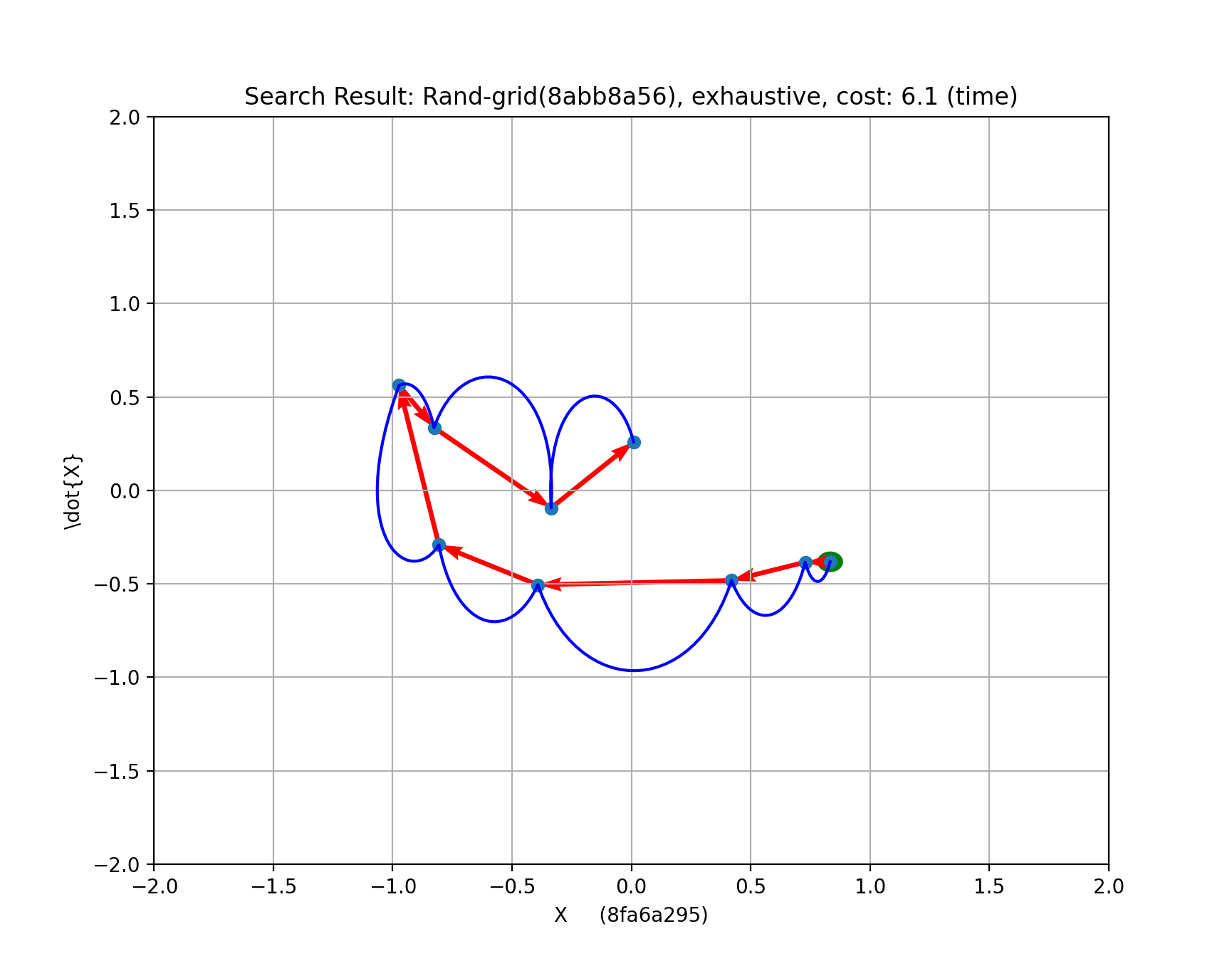}
  \includegraphics[width=\columnwidth]{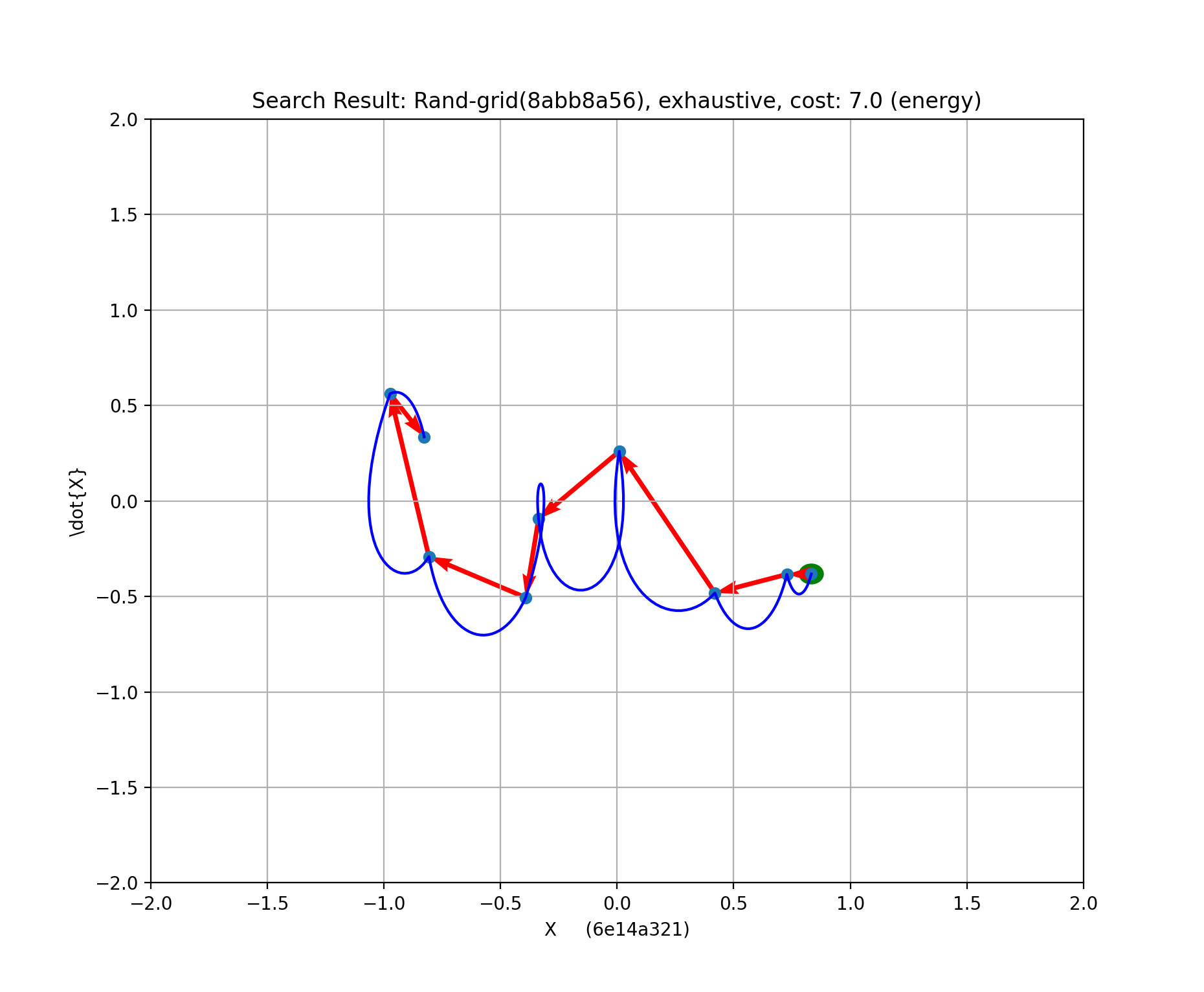}
  \caption{Globally optimum paths on the same  random 9 point grid
  as Figure \ref{3x3RandGridNN}, resulting from exhaustive search.
  Minimum  time cost path (Top) and minimum energy cost path (Bottom).
  Grid points used (8abb9a56).
  }\label{3x3RandGridBrute}
\end{figure}

%%%%** Section 3.2
\subsection{2D, $N=4$}
%%%%** Section 3.2.1 
\subsubsection{Rectangular Grid}

As computed above, exhaustive search for a global optimum with 2D, $N=4$ is
not feasible.   Instead we compared 1 million NN searches vs. 1 million
randomly sampled paths.

We generated one million random trajectories by shuffling the integers 0...15 and
evaluated their path costs with both time
and energy criteria on the rectangular grid.
The distributions of cost (Figure \ref{4x4x1Mdistribs}) are very close to Gaussian.  An apparent shift
between the Gaussian curve and the sampling bins in this figure
is an artifact of the bin
plotting. This was confirmed by generating 1M samples of synthetic data
with a known mean and observing a similar plot.

%%%%** Section 3.2.2 
\subsubsection{Normality}\label{SecNormality}
\begin{figure}\centering
  \includegraphics[width=\columnwidth]{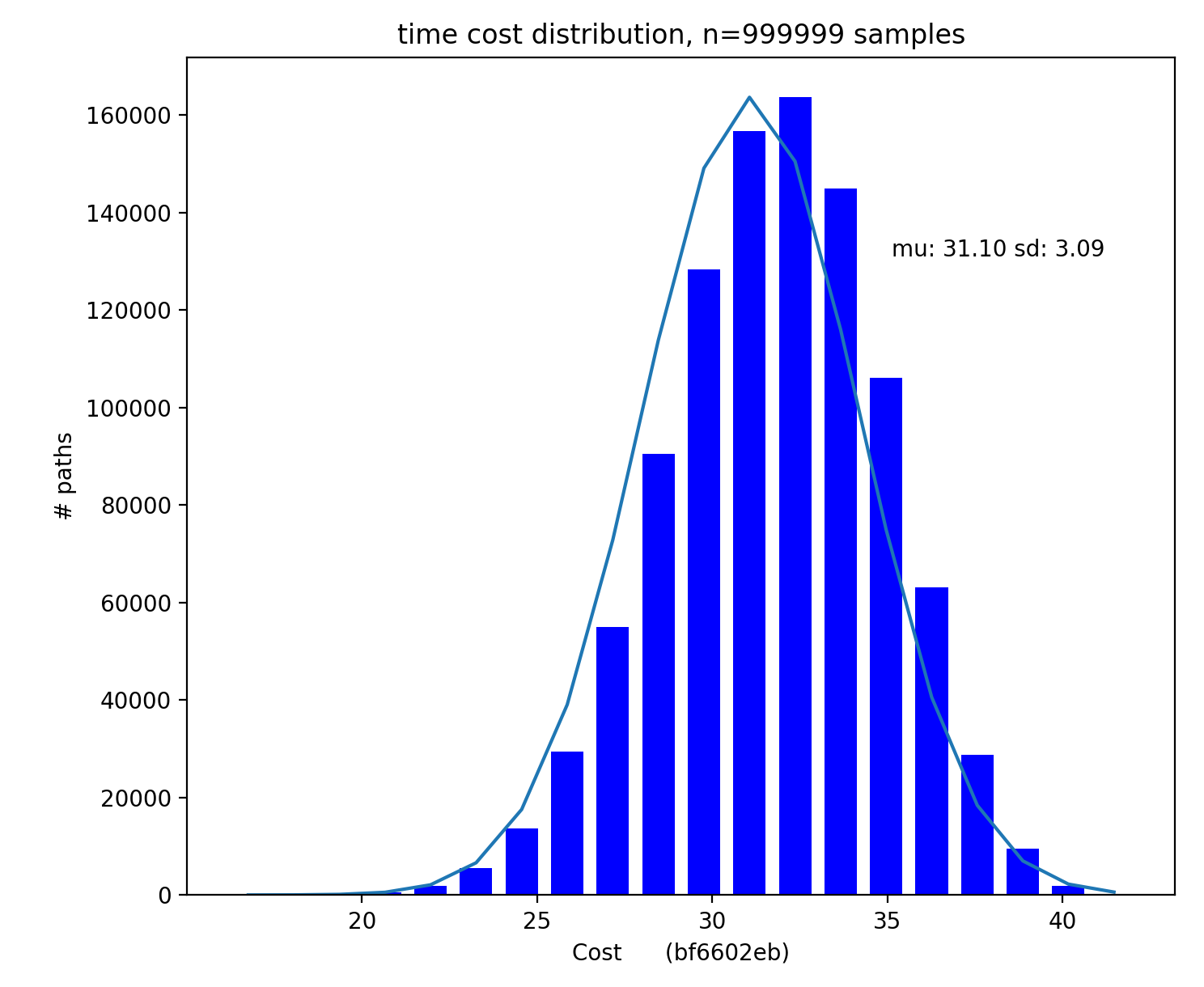}
  \includegraphics[width=\columnwidth]{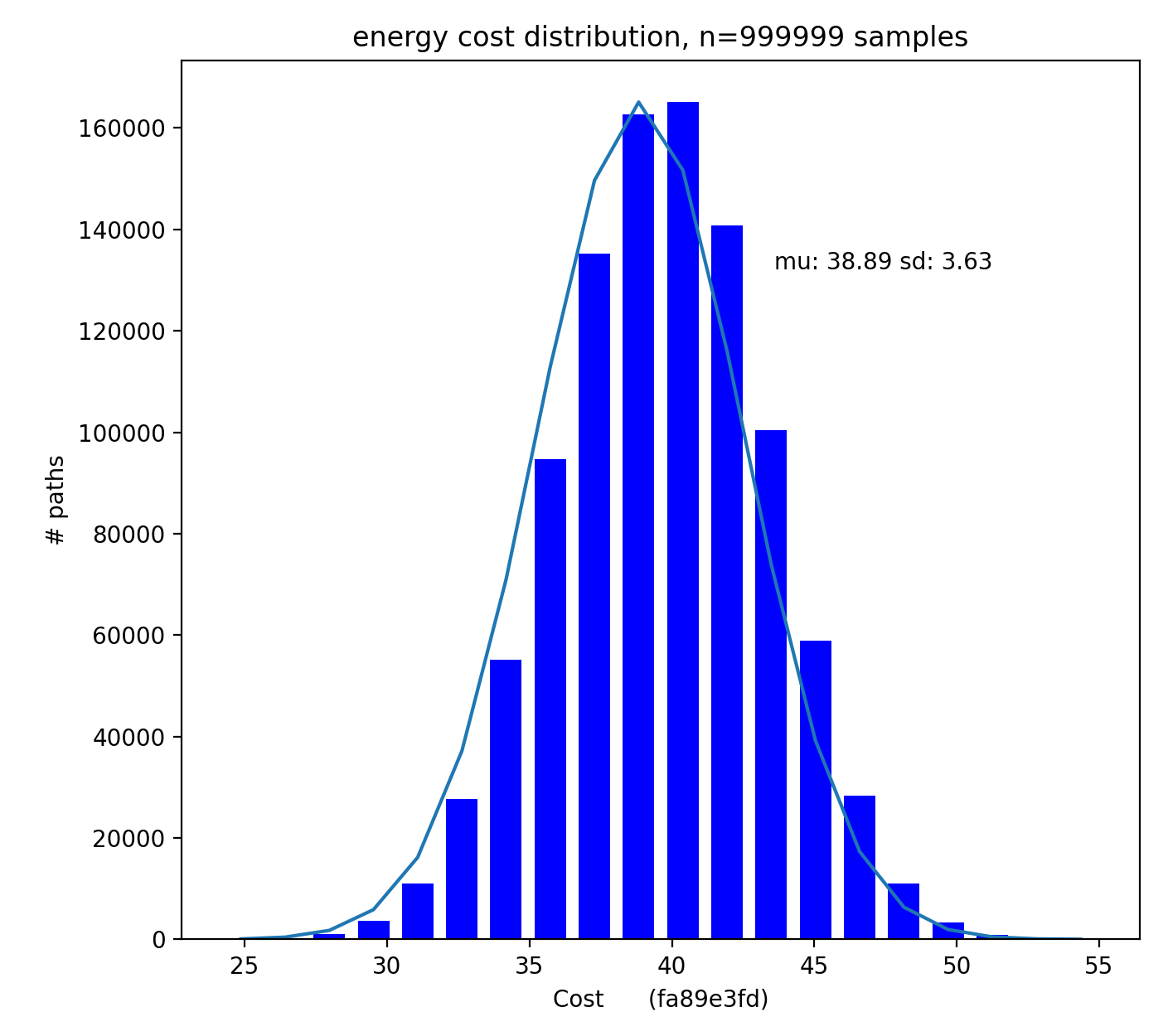}
  \caption{Distribution of one million randomly generated paths through the 2-D
  phase space (rectangular grid, $N=4$) by total time (Top)
  and total path energy use (Bottom).}\label{4x4x1Mdistribs}
\end{figure}

A Quantile-Quantile plot of 1M samples of
time cost from
Figure \ref{4x4x1Mdistribs}
(Figure \ref{QQ4x41M}) shows very close correspondence between the normalized
cost data with the normal distribution, $N(0,1)$.  Long duration outliers for
the time cost are sparse compared to Normal.

For evaluation of search results, the lower tail (negative quantiles) is of
most interest.  For time cost, the QQ plot extends below the $y=x$ line, indicating
that lower outliers are even rarer in the large sample
data than they are in the Normal distribution.
For energy cost, the situation is reversed, lower outliers are somewhat more common
in the large sample data than expected from the Normal distribution.

%%%%** Figure 5 

It is noteworthy that the minimum time cost of the 1M sample trajectories
(Figure \ref{4x4x1Mdistribs})
was approximately 16.1 which is quite far out on the left tail of the
sample's distribution.
For this particular sample of one million paths,
the   best time is $4.85\sigma$ below the mean.
The minimum energy cost from the sample trajectories was 23.1 which is
$4.4\sigma$ below the mean.  We compare these outliers discovered by
random searching to the NN results in Section \ref{SecSampleStats} below.

%%%%** Section 3.3
\subsection{Random Grid}\label{randomPts2D}

We repeated the computations of Section \ref{3x3RectSection}
with a 2D random grid of 16 points ($N=4$).

%%  Fig '1r'

\begin{figure}\centering
\centering
\begin{subfigure}{0.35\textwidth}
    \includegraphics[width=\textwidth]{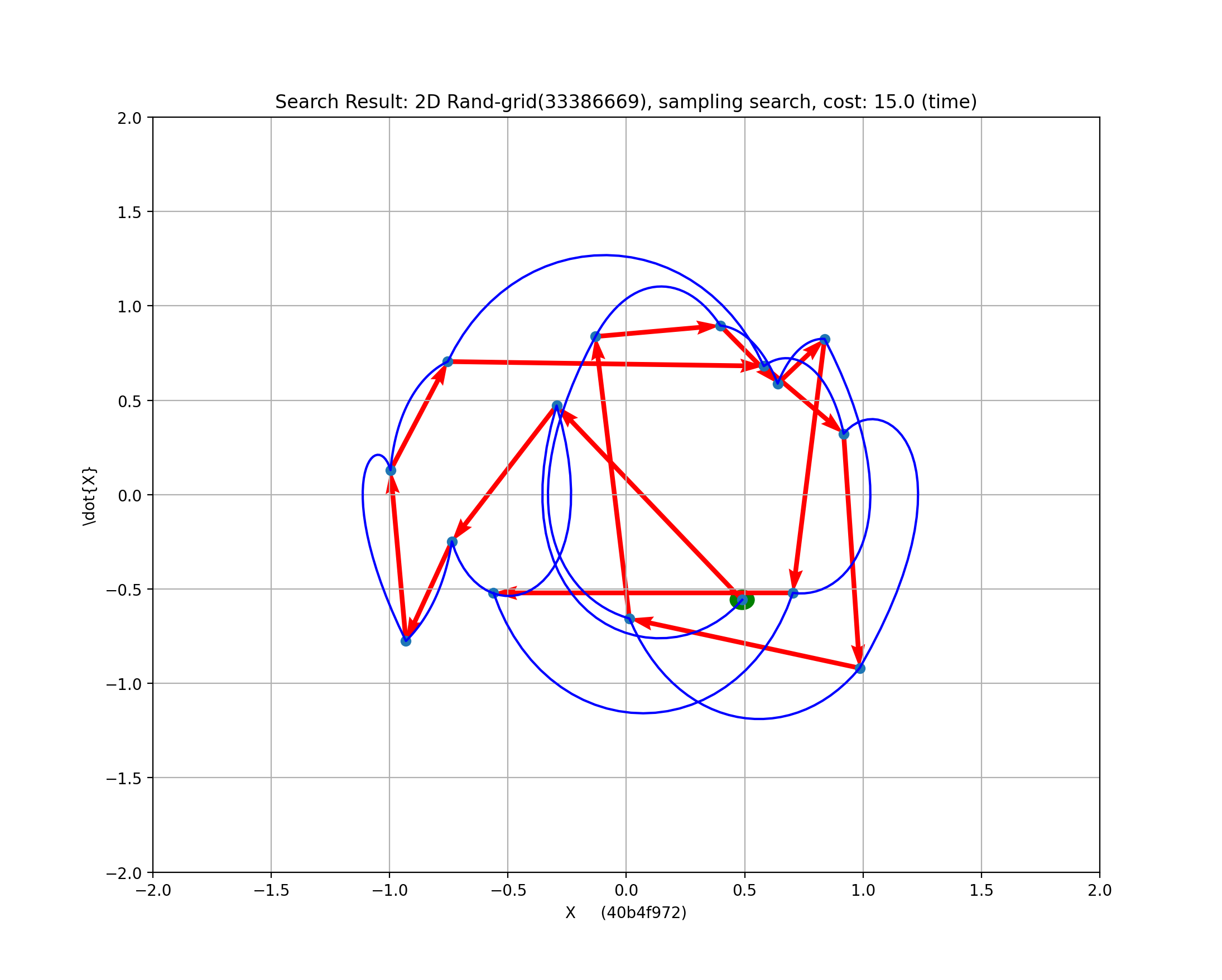}
    \caption{Suboptimal: Time Cost}
    \label{fig7:first}
\end{subfigure}
\hfill
\begin{subfigure}{0.35\textwidth}
    \includegraphics[width=\textwidth]{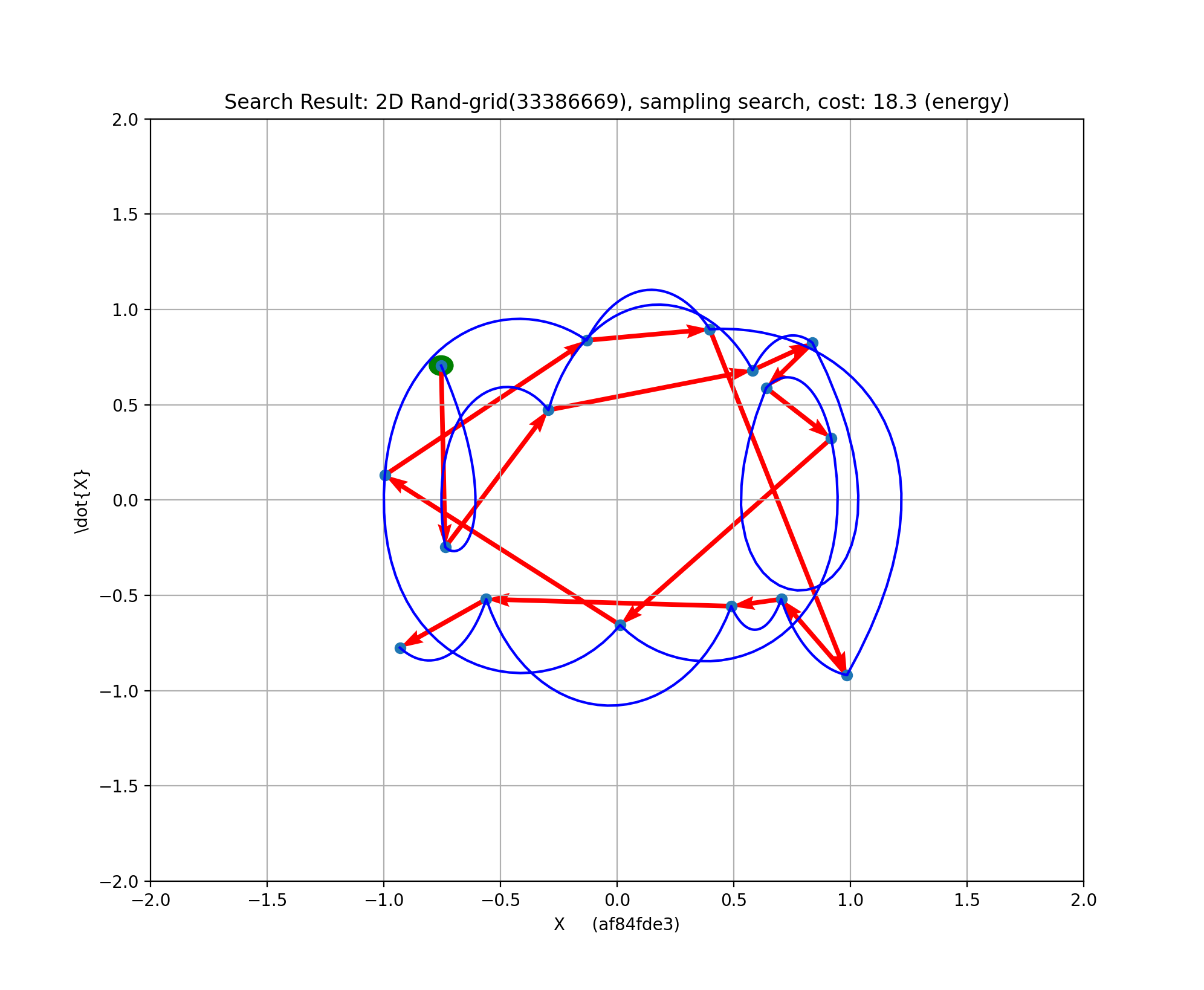}
    \caption{Suboptimal: Energy Cost}
    \label{fig7:second}
\end{subfigure}
\hfill
\begin{subfigure}{0.35\textwidth}
    \includegraphics[width=\textwidth]{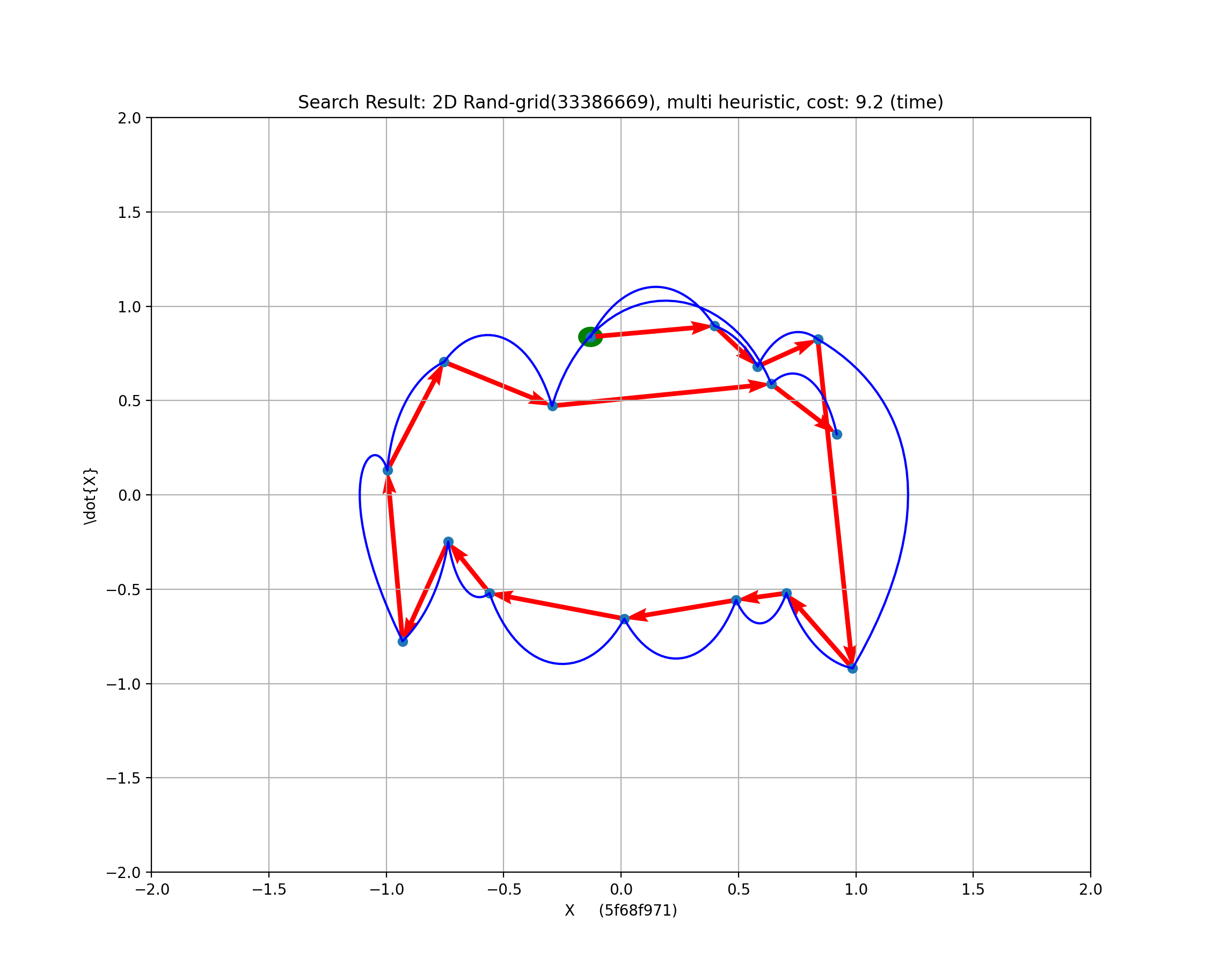}
    \caption{Globally optimal: Time Cost}
    \label{fig7:third}
\end{subfigure}
\begin{subfigure}{0.35\textwidth}
    \includegraphics[width=\textwidth]{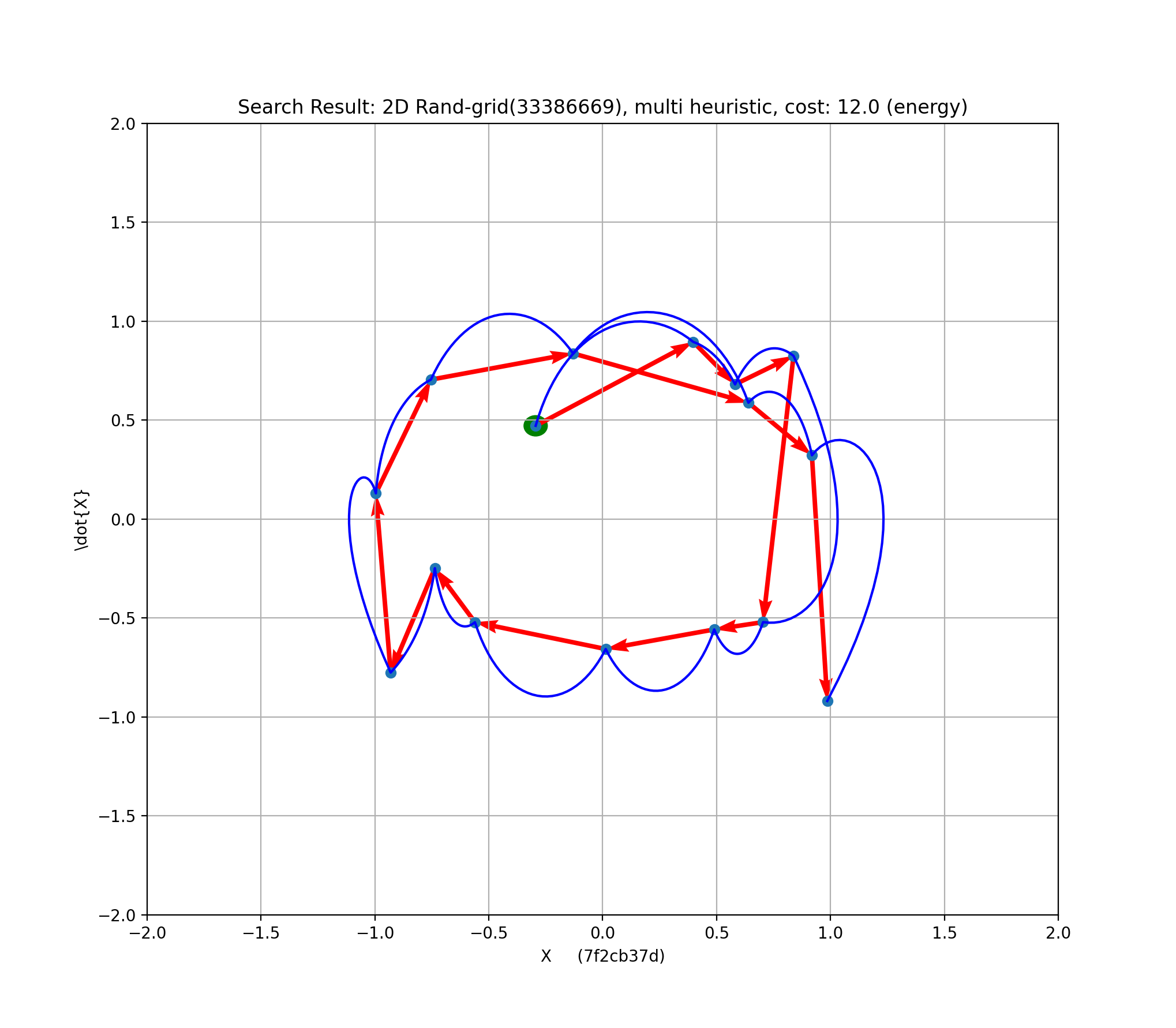}
    \caption{Globally optimal: Energy Cost}
    \label{fig7:fourth}
\end{subfigure}
  \caption{ 2D, $N=4$ random grid.
  Best paths from 50,000 random sample paths: (a) total time (b)
  energy use.
  Best paths from 50,000 NN searches (c, time, d, energy).
  (random grid: 33386669).  }  \label{RandGrid2x4trajectories}

\end{figure}

First we illustrate the best trajectories found with 50,000 randomly sampled trajectories
to the best trajectories with 50,000 Multi-Heuristic searches (Figure \ref{RandGrid2x4trajectories}).
Recall that we could not find the global optimum for the 2D, $N=4$ grid due to
extremely long required computation time.  The more optimal paths (lower row of
Figure \ref{RandGrid2x4trajectories}) are simpler and pursue clockwise trips through the points.
Costs for the trajectories illustrated were substantially lower than the best
random sampled trajectories (time: 9.2 vs. 15.0,
energy, 12.0 vs. 18.3).  These overall effects were similar to the regular grid (Figure
\ref{3x3ExhaustivRectGrid}).

%% Fig '2r'

%%%%** Figure 7 

\begin{figure}\centering   % 9/4/23
  \includegraphics[width=\columnwidth]{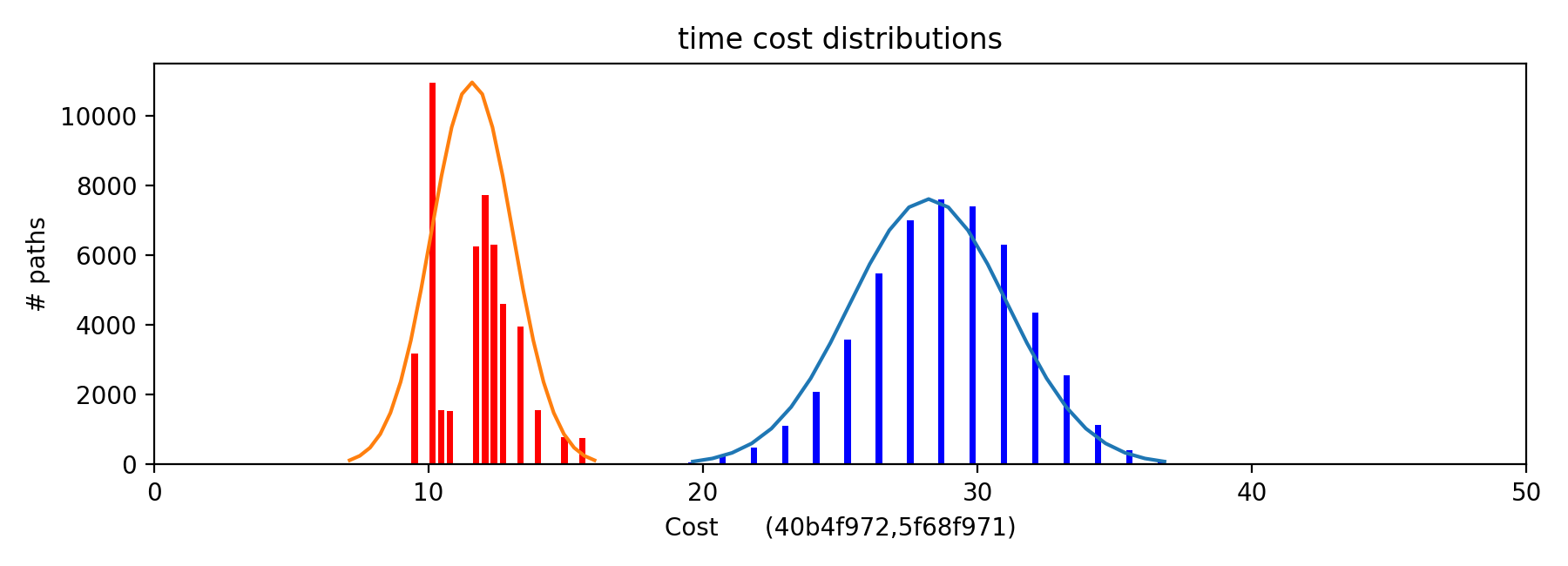}
  \includegraphics[width=\columnwidth]{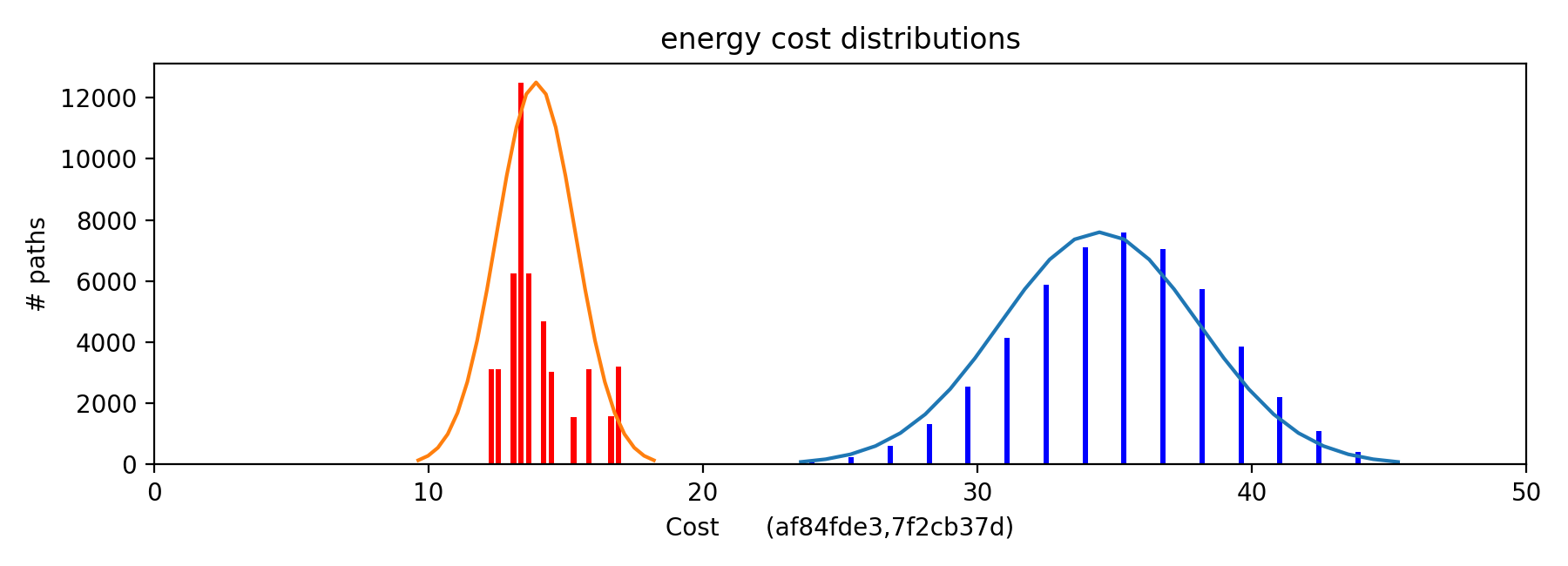}
  \caption{Comparing distributions of 50,000 nearest-neighbor
  heuristic (NN) paths (red) with 50,000 random paths (all 16
  starting points) through the 2-D   grid ($N=4$) by total time (Top)
  and total path energy use (Bottom).(random grid: 33386669).}\label{RandGrid2x4CostDiff}
\end{figure}

Comparing the NN cost distributions with the sampled distributions for the
random grid (Figure \ref{RandGrid2x4CostDiff}) shows   little or no overlap
between the distributions, and the NN distribution (red), while still non-Gaussian,
is becoming closer.

%%%%** Figure 8 

\begin{figure}\centering   % 9/4/23
  \includegraphics[width=\columnwidth]{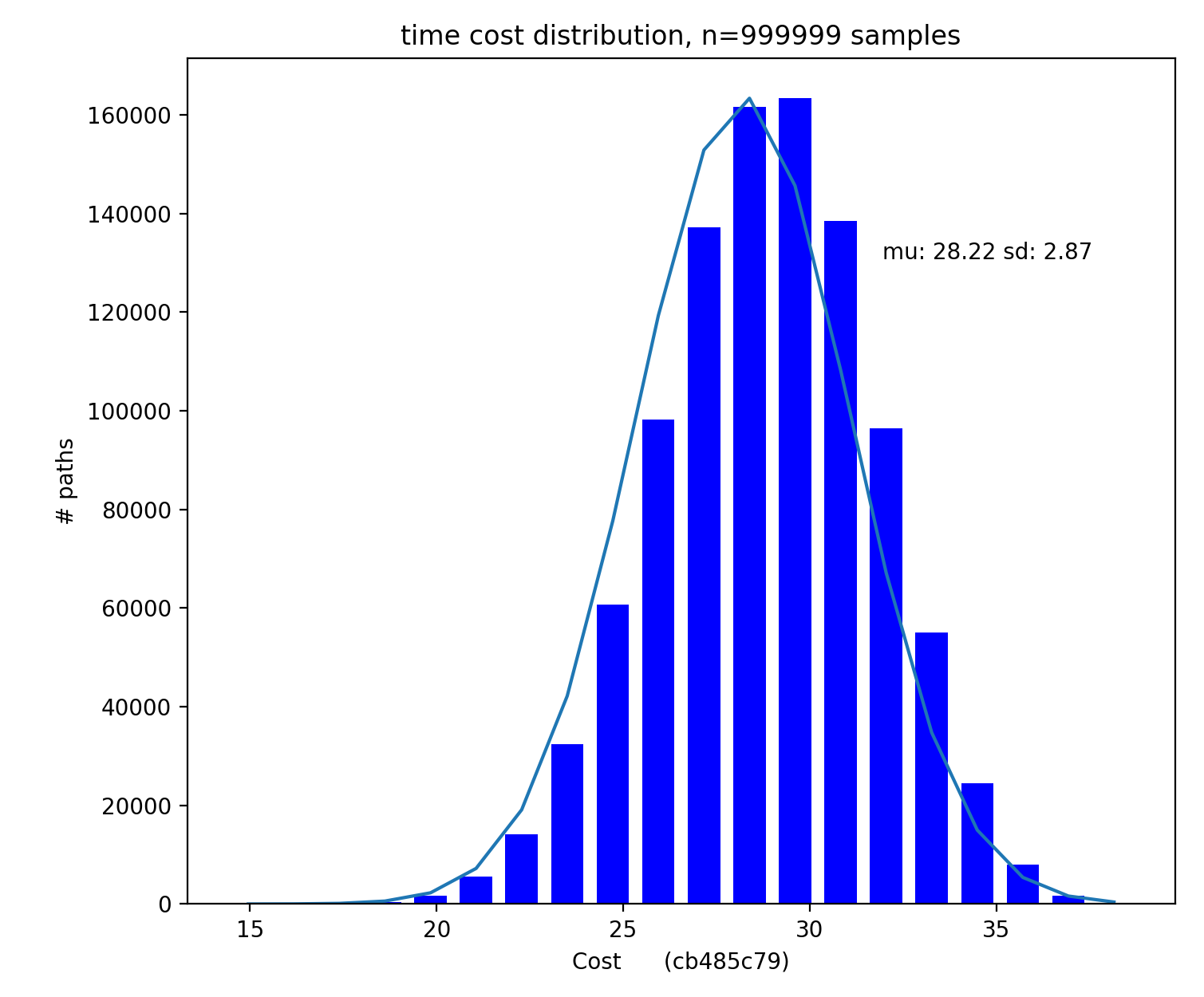}
  \includegraphics[width=\columnwidth]{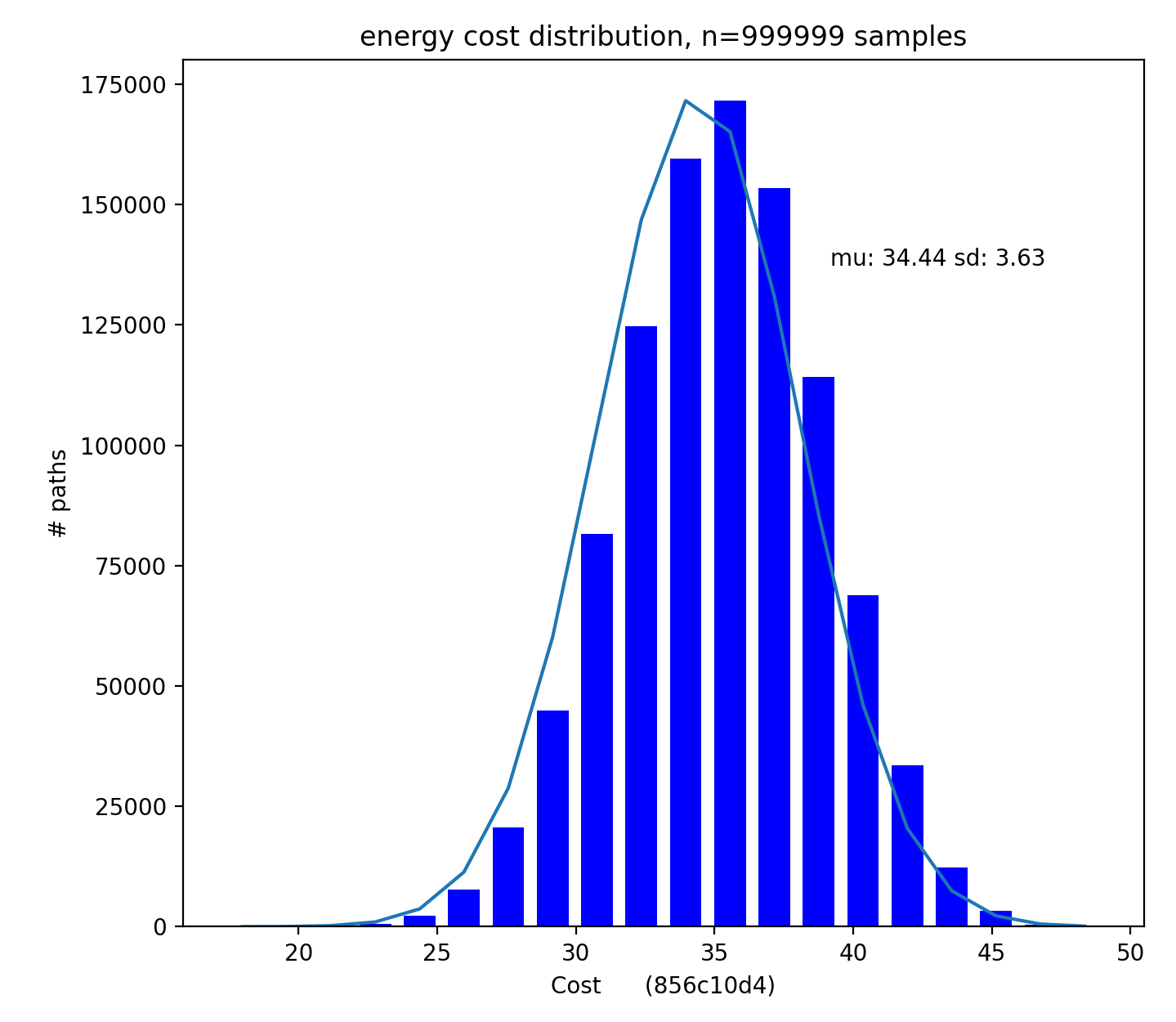}
  \caption{Distribution of one million randomly generated paths through a 2-D,
  $N=4$ random grid
  phase space   by total time (Top)
  and total path energy use (Bottom).
  (random grid ID: 33386669).}\label{RandGrid2x4x1Mdistribs}
\end{figure}

\begin{figure}\centering
  \includegraphics[width=\columnwidth]{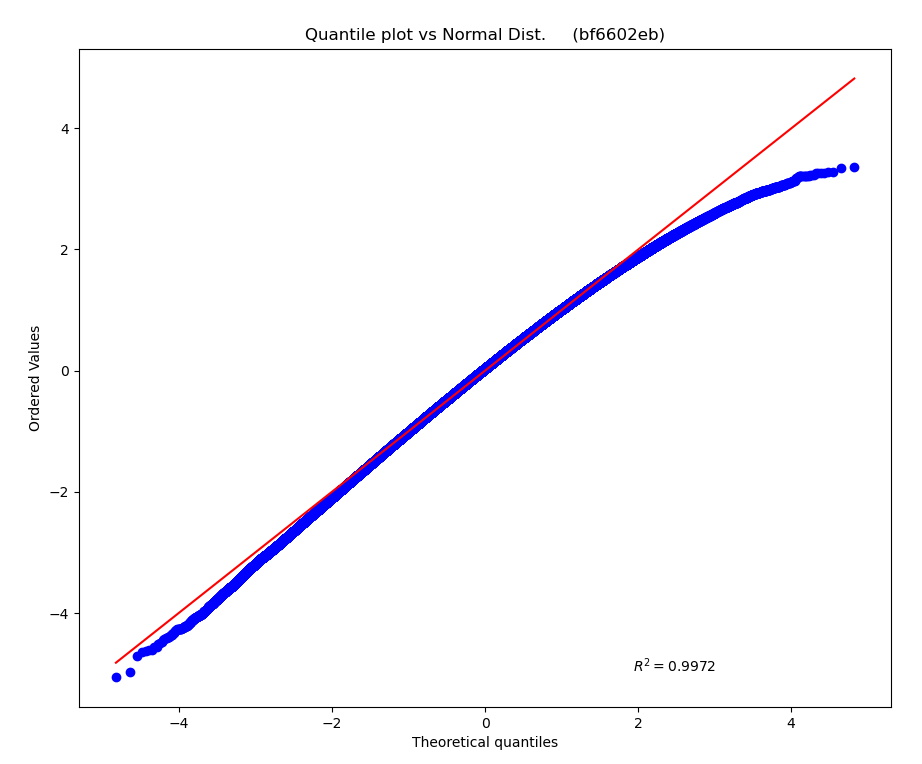}
  \includegraphics[width=\columnwidth]{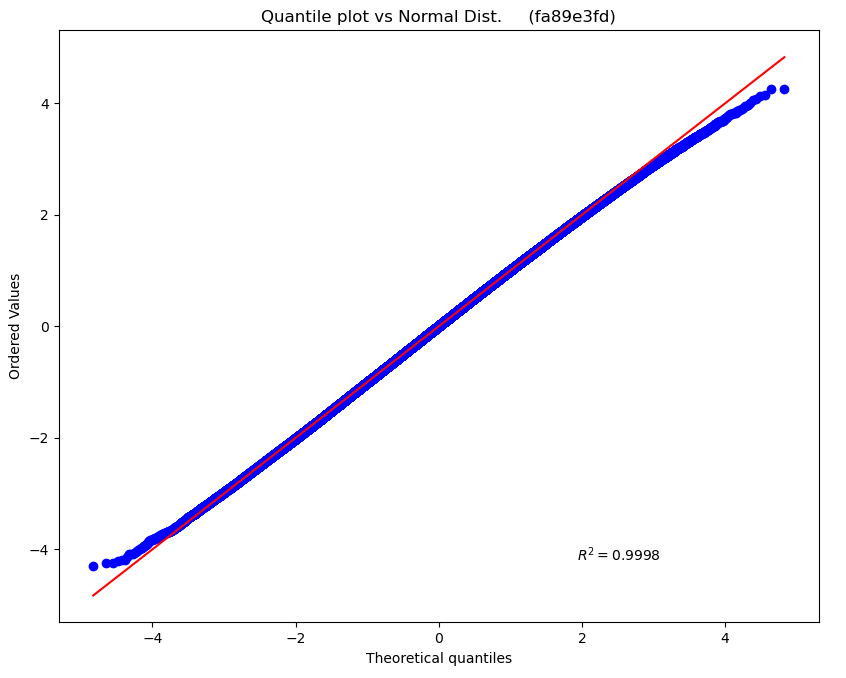}
  \caption{Quantile plots of costs from searching 1M points {\bf 2D, 4x4 Rectangular} grid (Fig.\ref{4x4x1Mdistribs})
  compared with normal distribution (after normalization to mean and standard
  deviation). Time cost (Top), energy cost (Bottom).}\label{QQ4x41M}
\end{figure}

\begin{figure}\centering   % 9/4/23
  \includegraphics[width=\columnwidth]{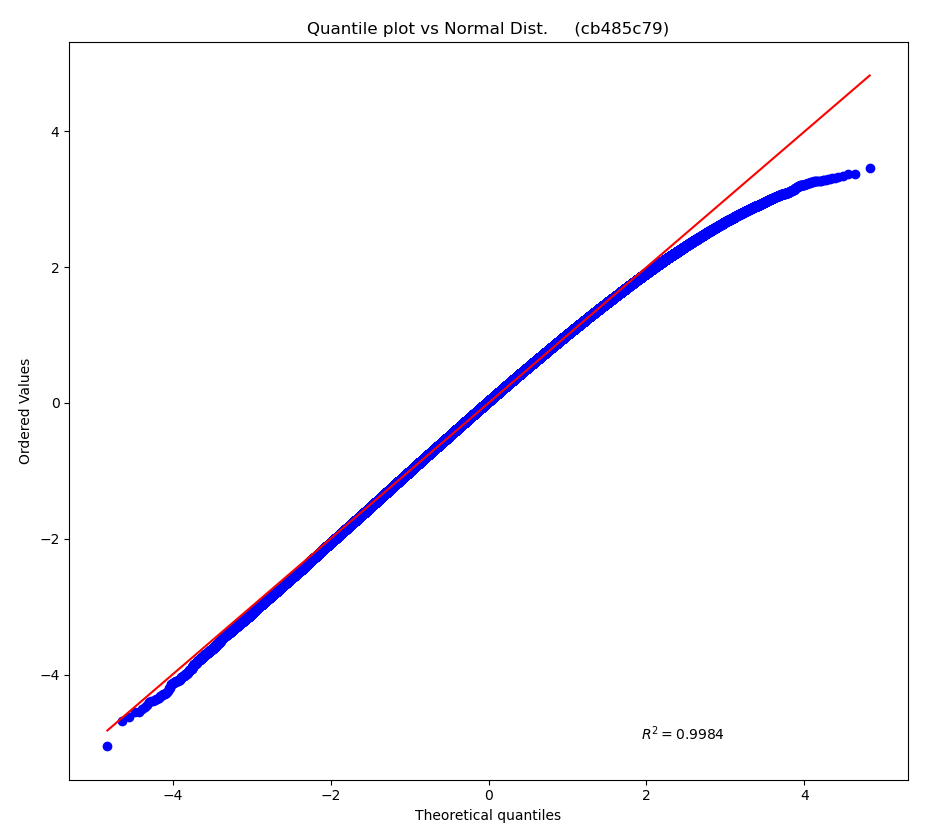}
  \includegraphics[width=\columnwidth]{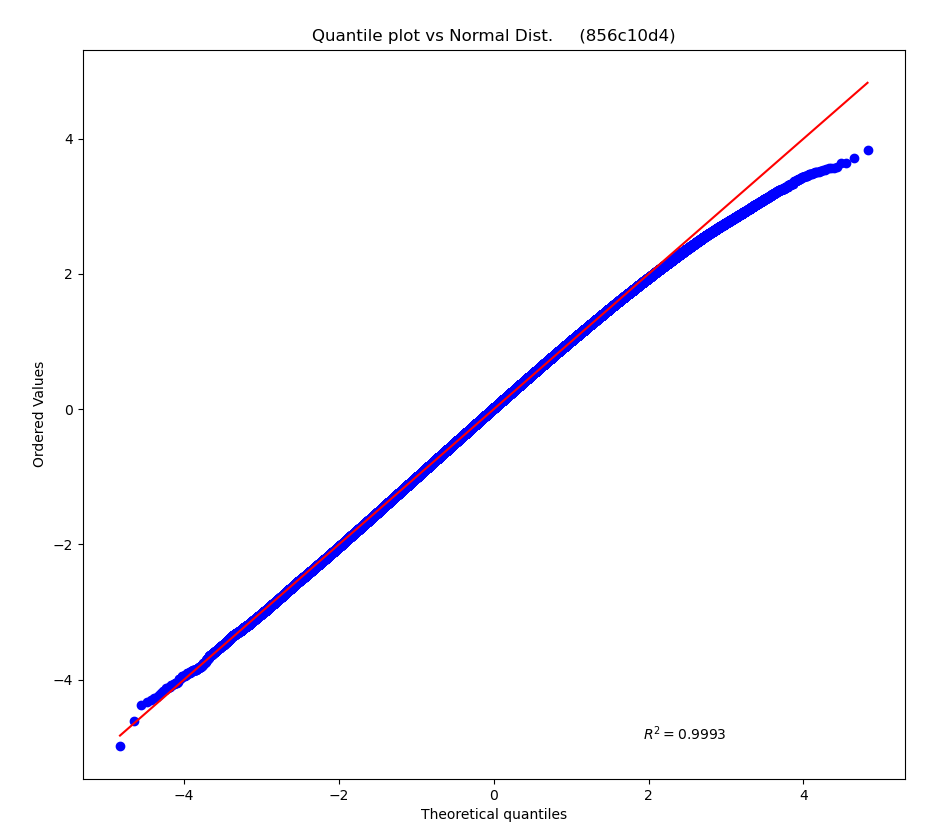}
  \caption{Costs from 1M searches of {\bf 2D, 16, Random} grid (Fig.
  \ref{RandGrid2x4x1Mdistribs})/
  compared with normal distribution (after normalization to mean and standard
  deviation). Notably, the negative tail is closer to a perfect normal distribution
  with the random grid. Time cost (Top), energy cost (Bottom).
  (random grid: 33386669)}\label{RG_QQ4x41M}
\end{figure}

We again assessed the degree that 1M samples of trajectories on the 2D, $N=4$ random grid
fit a normal distribution using the histograms (Figure \ref{RandGrid2x4x1Mdistribs}
and the quantile-quantile plot (Figure \ref{RG_QQ4x41M}).
The substitution of a random grid for the
rectangular grid seems to improve the closeness of fit to a Gaussian, particularly in the
important lower tail (compare Figure \ref{RG_QQ4x41M} with Figure \ref{QQ4x41M}).

%% Fig '5r'
%%%%** Figure 9 

%% Fig 6r

\afterpage{\clearpage}

%%%%** Figure 6

%%%%** Figure 10 

%%%%** Section 3.4
\subsection{6D}

We now explore the 6D space arising from considering the $X,Y,Z$ positions in
a 3D work volume crossed with the end-effector  velocities, $\dot{X}, \dot{Y}, \dot{Z}$.

%%%%** Section 3.4.1 
\subsubsection{Rectangular Grid}

\begin{figure}\centering      %%  REGULAR GRID
  \includegraphics[width=\columnwidth]{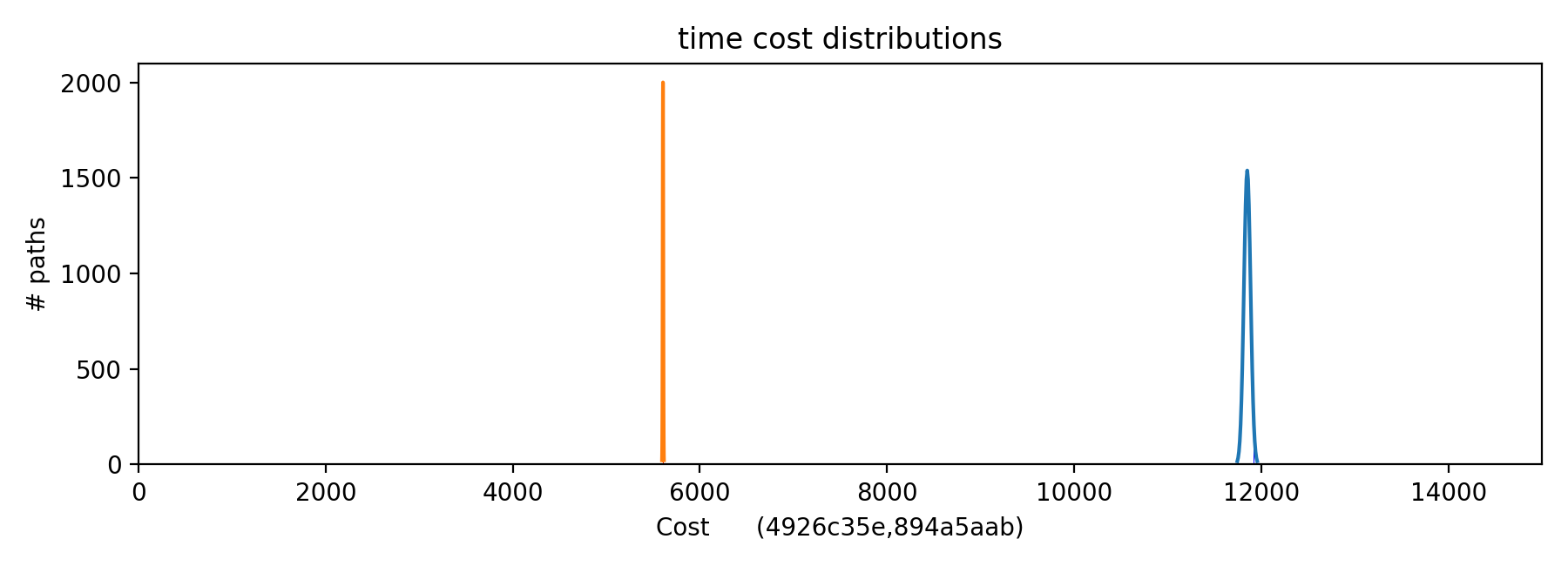}
  \includegraphics[width=\columnwidth]{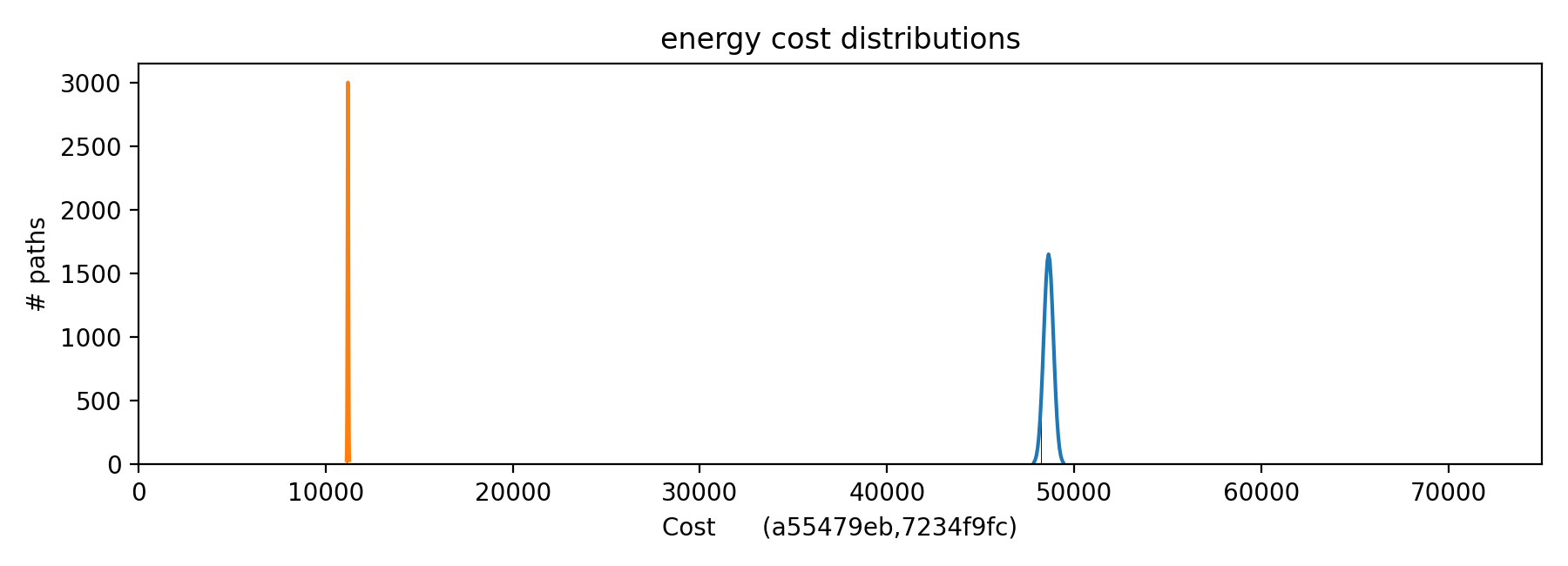}
  \caption{Distributions for time cost (Top) and energy cost (Bottom) of the
  6D 4x4 rectangular grid.
  Note scale change on Energy cost graph.
  10,000 random samples (Blue) and
  10 nearest neighbor heuristic (NN) searches (Red). Magnitude of
  NN search result scaled by area to visually match random samples.}\label{RectangularGridTwoWayPlots}
\end{figure}

 We ran 10,000 NN searches
in the 6D, 4x4 rectangular grid, and compared them to a sample of 10,000 random
paths (Figure \ref{RectangularGridTwoWayPlots}). Cost reduction was dramatic for
both time and energy costs.
However, each 6D, $N=4$ NN search took about 1000 times
as much computation time as evaluating the cost of a random path.  See Section \ref{SecSampleStats} for consideration of this tradeoff.

\afterpage{\clearpage}

%%%%** Figure 13
\begin{figure}\centering   % 4/10/24
  \includegraphics[width=0.97\columnwidth]{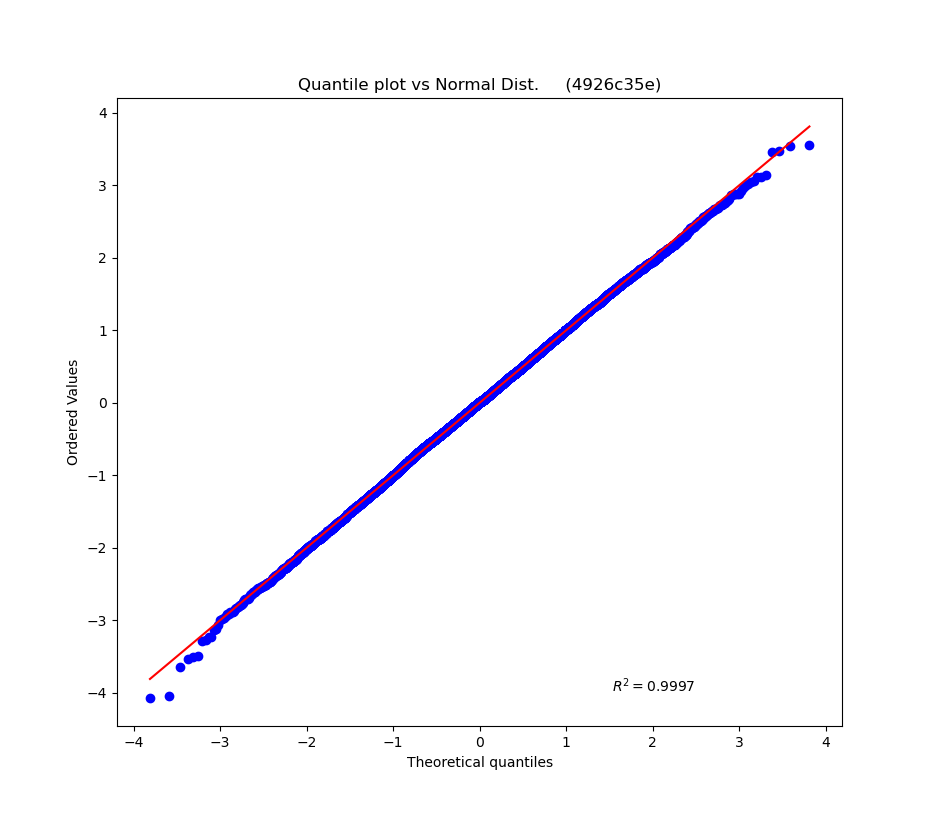}
  \includegraphics[width=0.97\columnwidth]{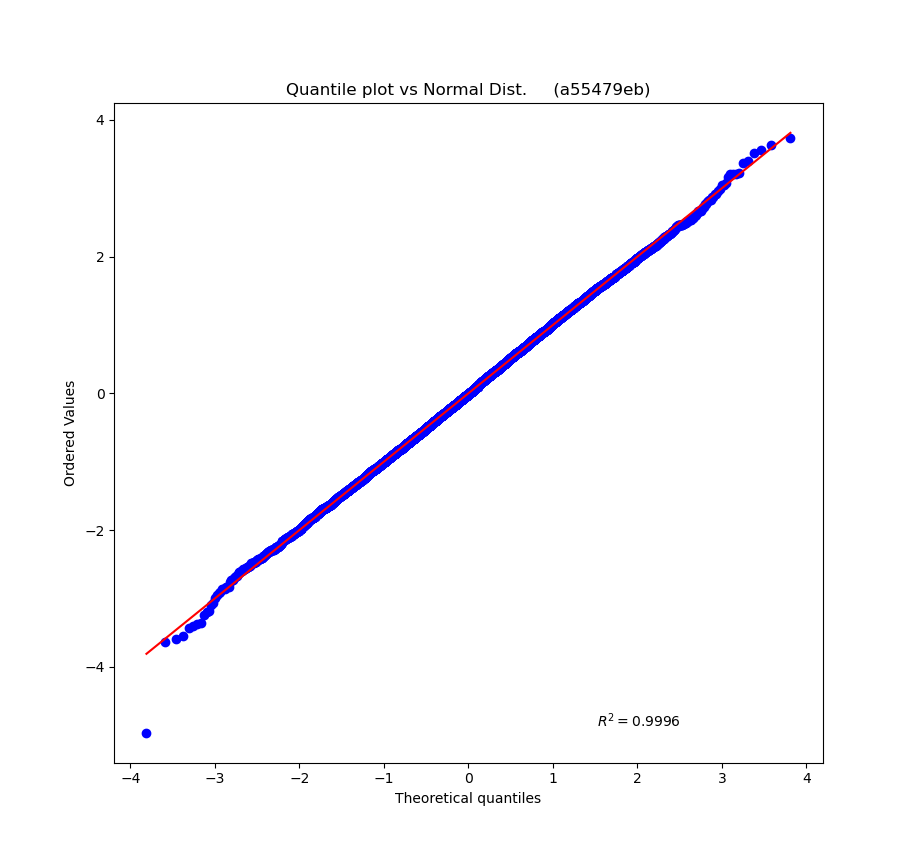}
  \caption{Normality with the {\bf 6Dx4x4 Rectangular} grid.   Quantile plots of 10,000 random trajectories from Figure \ref{RectangularGridTwoWayPlots}
  (blue curve)
  compared with normal distribution (after normalization to mean and standard
  deviation). Time cost (Top), energy cost (Bottom). except for a very low cost outlier for Energy cost. }
  \label{6Dx4x4RectQQ}
\end{figure}

%%%%** Figure 14
\begin{figure}\centering   % 4/10/24
  \includegraphics[width=\columnwidth]{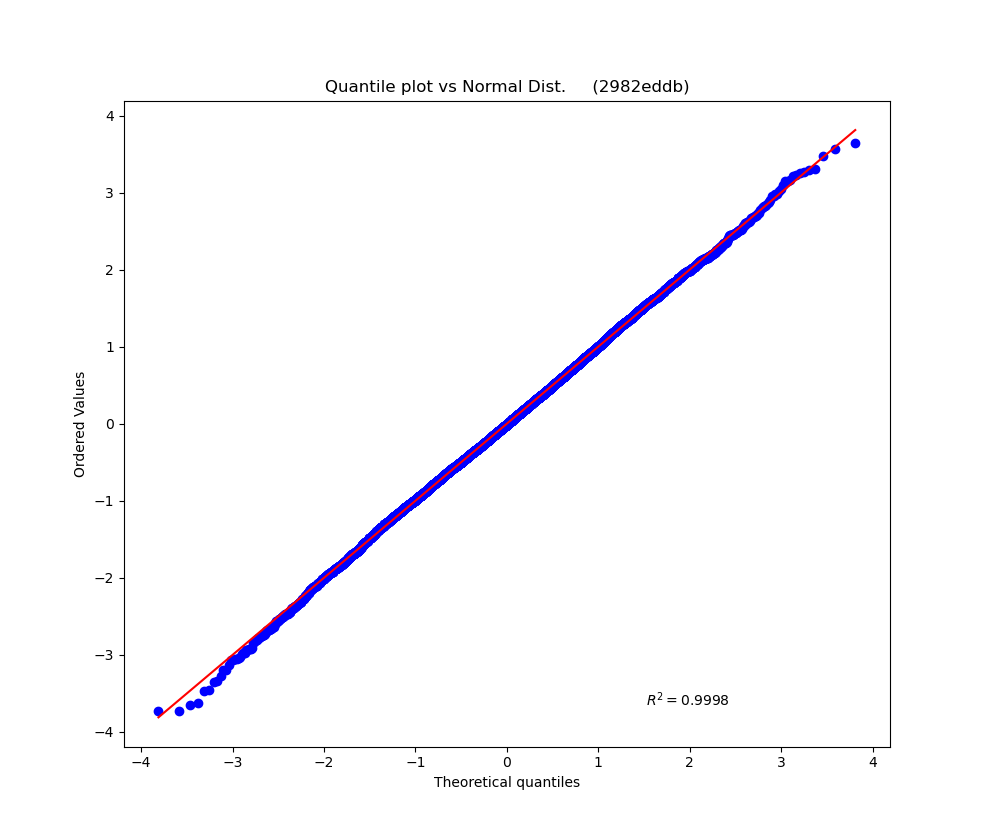}
  \includegraphics[width=\columnwidth]{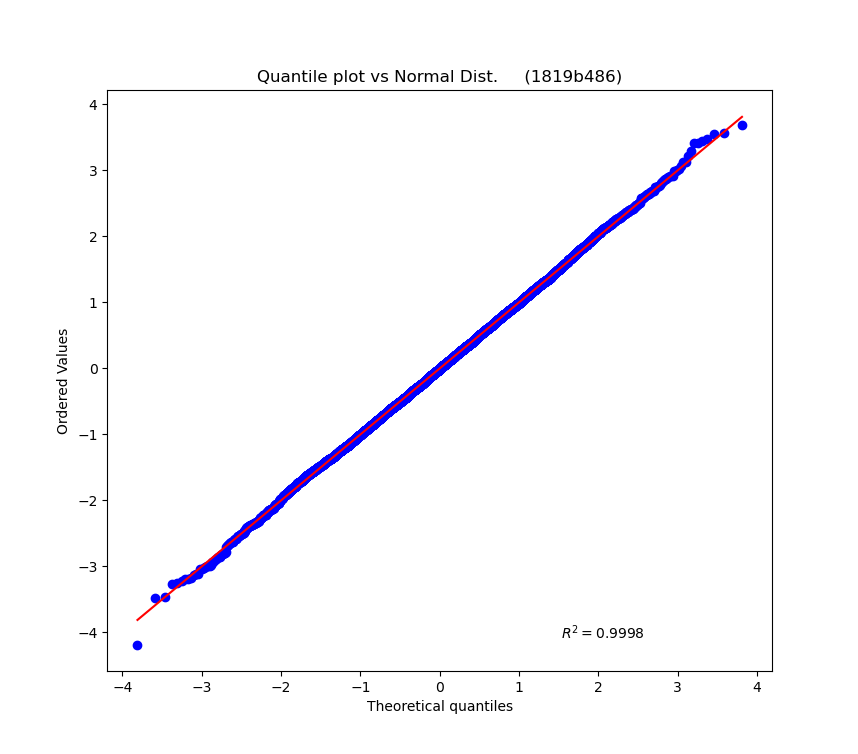}
  \caption{Normality with the {\bf 6Dx16 Random grid}.  Quantile plots of 10,000 random trajectories from Figure \ref{RandGridTwoWayPlots}
  (blue curve)
  compared with normal distribution (after normalization to mean and standard
  deviation). Notably, the negative tail is closer to a perfect normal distribution
  with the random grid (compare with Fig. \ref{RectangularGridTwoWayPlots}).
  Time cost (Top), energy cost (Bottom).
  (random grid: 165732fc)}\label{6Dx4x4RANDOMQQ}
\end{figure}

\afterpage{\clearpage}

%%%%** Figure 11 

%%%%** Figure 12 
 \begin{figure}\centering   %
  \includegraphics[width=\columnwidth]{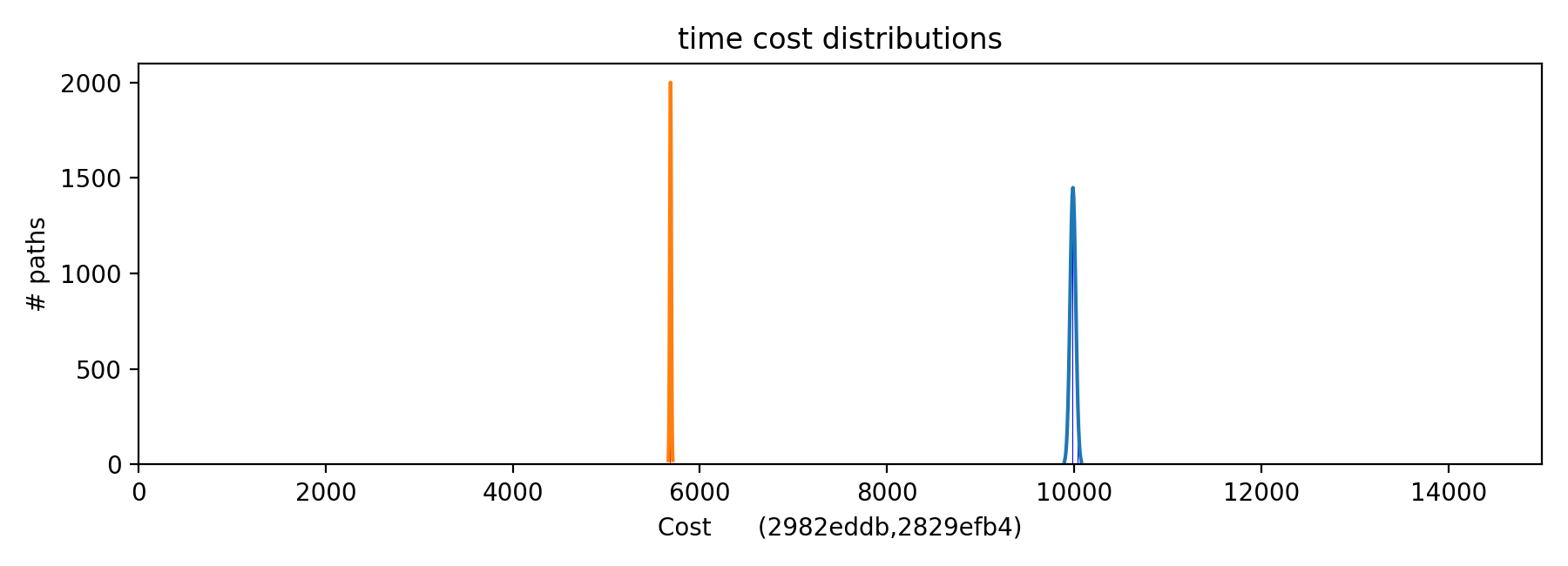} %corrected 8/31
  \includegraphics[width=\columnwidth]{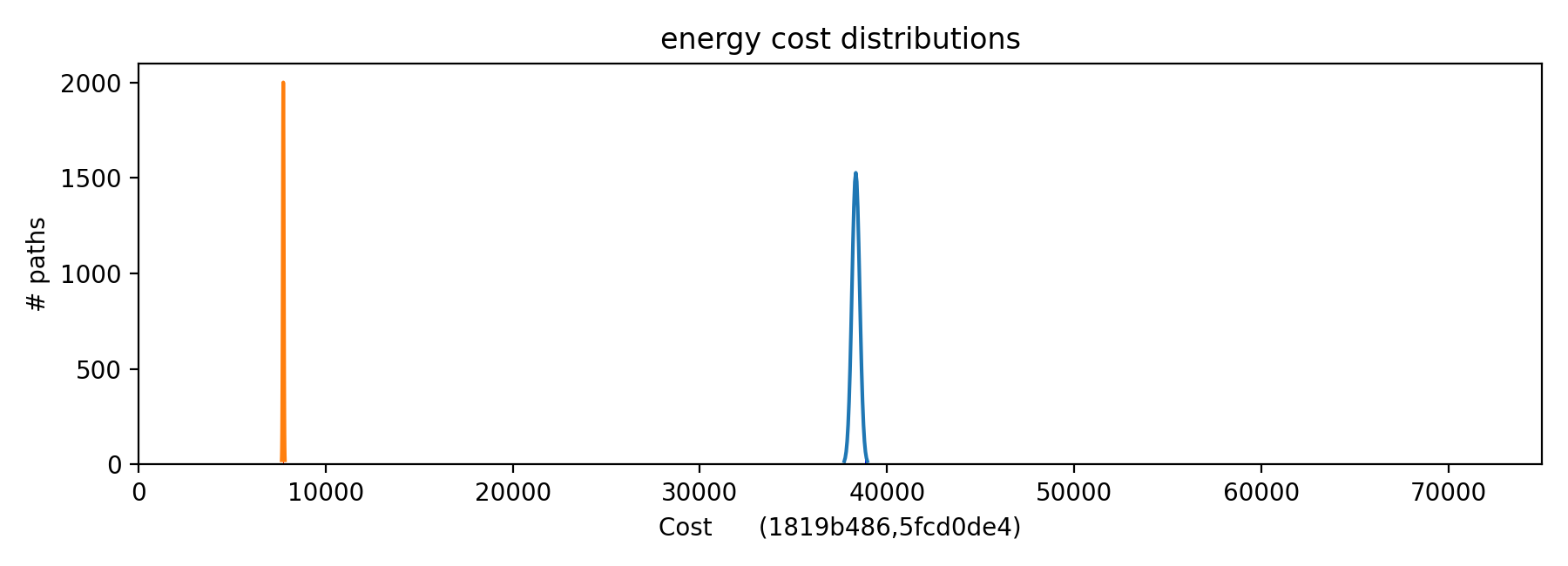}
  \caption{Cost comparisons with a 6D, $N=4$, random instead of rectangular grid for time cost (Top) and energy cost (Bottom).
  Random samples (Blue) and  NN searches (Red).
  (points dataset 165732fc).}\label{RandGridTwoWayPlots}
\end{figure}

%%%%** Figure 14
\begin{figure}[h]\centering   % New data 8/31
\begin{subfigure}{0.37\textwidth}
    \includegraphics[width=\textwidth]{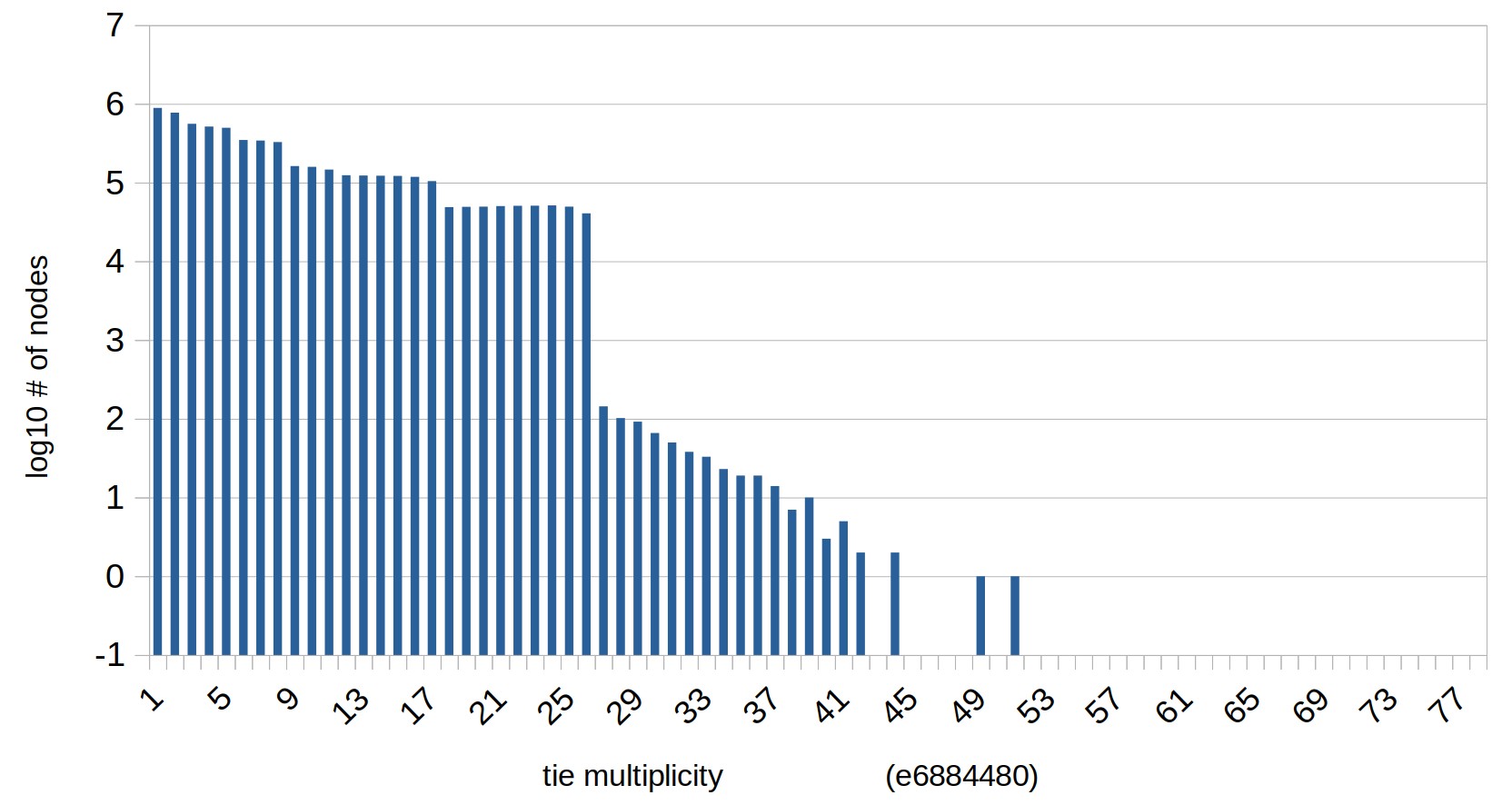}
    \caption{Rectangular Grid: Time Cost}
    \label{fig14:first}
\end{subfigure}
\hfill
\begin{subfigure}{0.37\textwidth}
    \includegraphics[width=\textwidth]{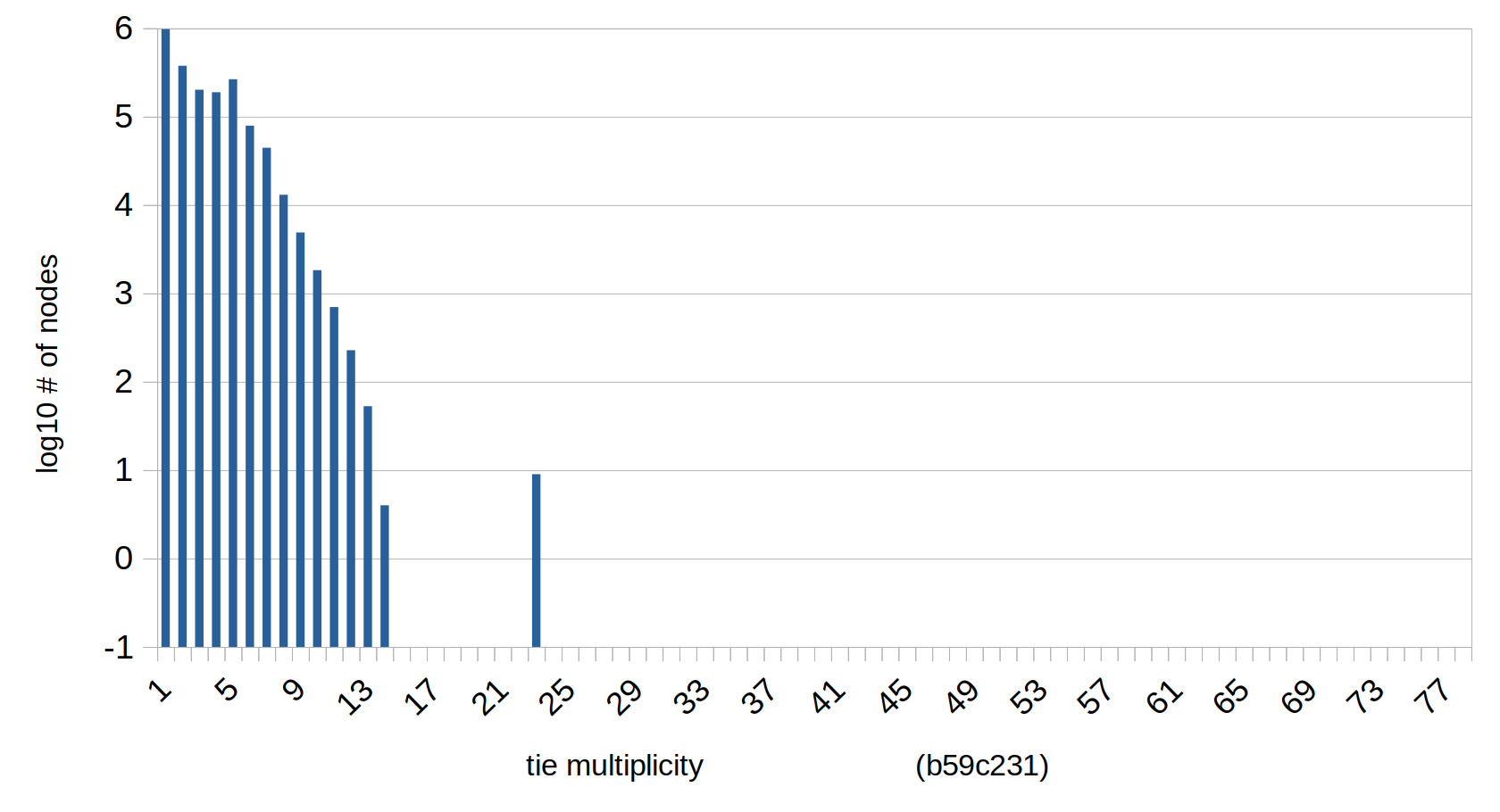}
    \caption{Rectangular Grid: Energy Cost}
    \label{fig14:second}
\end{subfigure}
\hfill
\begin{subfigure}{0.37\textwidth}
    \includegraphics[width=\textwidth]{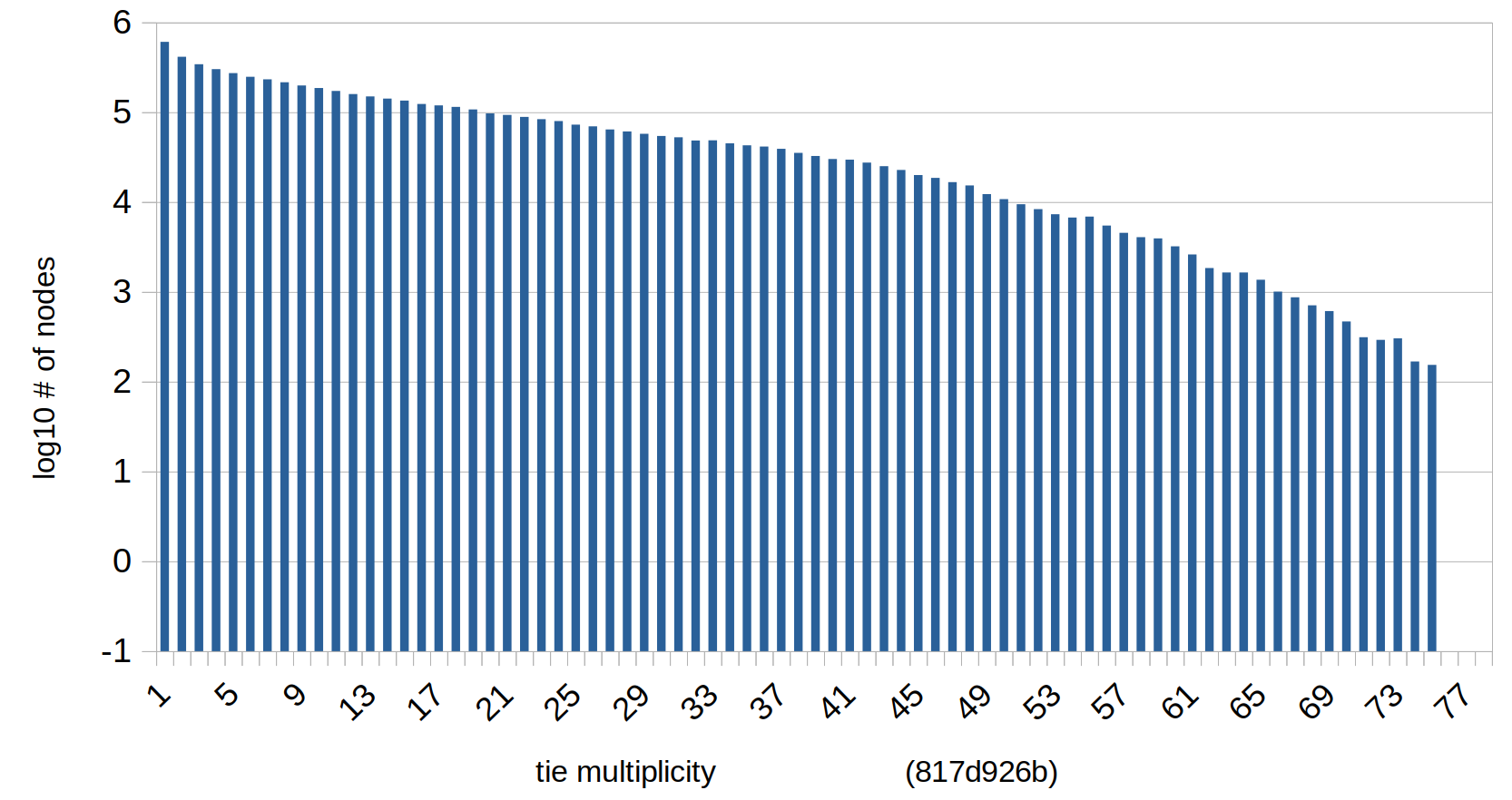}
    \caption{Random Grid: Time Cost}
    \label{fig14:third}
\end{subfigure}
\begin{subfigure}{0.37\textwidth}
    \includegraphics[width=\textwidth]{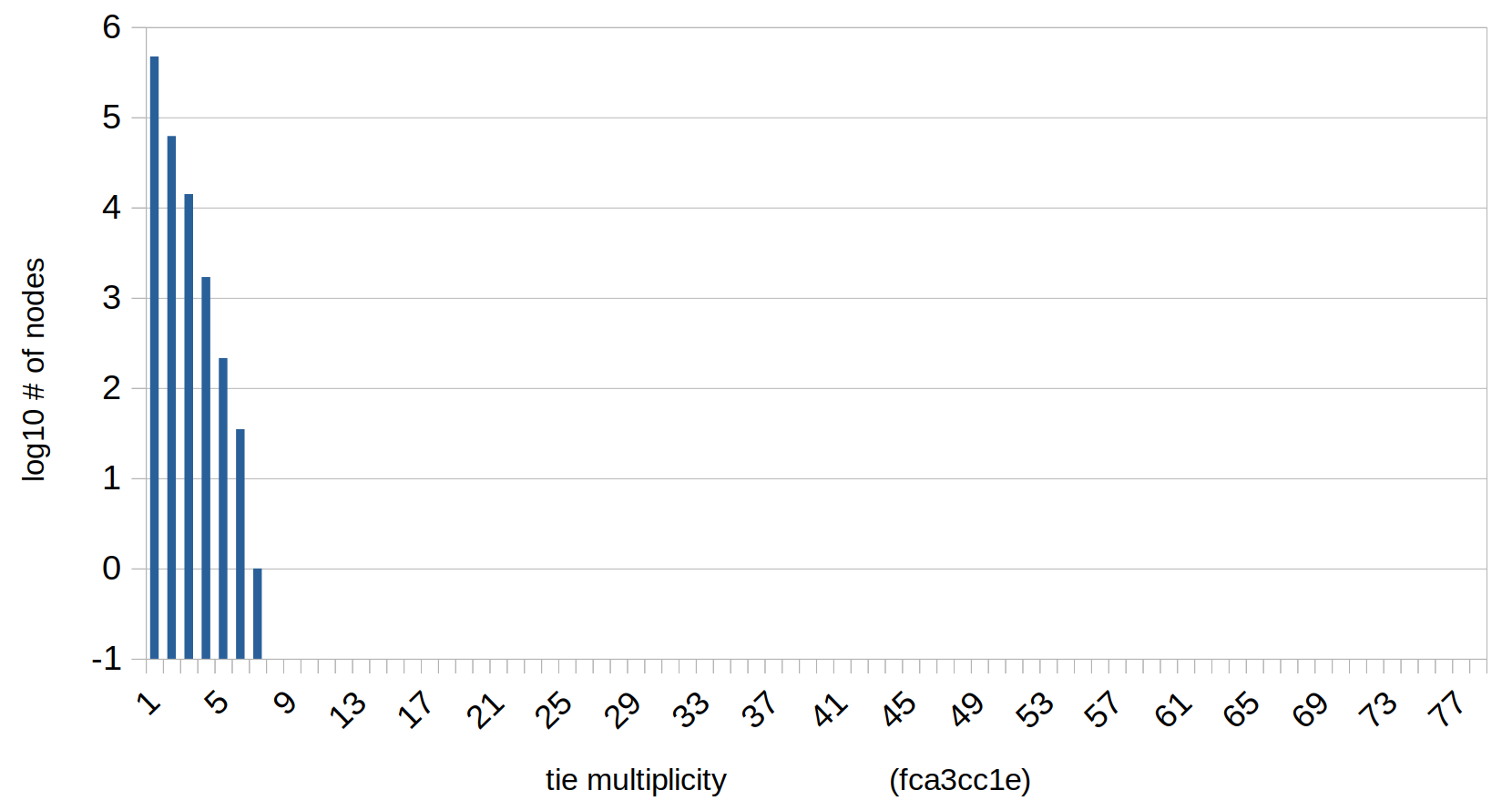}
    \caption{Random Grid: Energy Cost}
    \label{fig14:fourth}
\end{subfigure}
\caption{Distribution of the log number of ties during 10,000 NN heuristic searches
6D, $N=3$ space (~14 searches per starting point, random point set 73d954cc).
For Time cost on random grid (c),
the apparent truncation of the distribution at about
76 branches is in the data itself and not due to plotting axis choice.}\label{tiesTimeEnergy}\label{RandPointTiesTimeEnergy}
\end{figure}

%%%%** Section 3.4.2 
\subsubsection{Random Grid}

As with the 2D random grid, 6D points were generated at random from a uniform distribution
such that for the six dimensions, each coordinate was an independent random variable
between $-1$ and $1$.

 With all points defined randomly, we ran 10 NN searches for
comparison with 10,000 random sampled paths using both the time and energy costs.
Execution time for these two computations was about the same as for the
rectangular grid (15-20 min.).

\paragraph{Cost performance on the random points}
A sample of random paths had lower time costs with randomized points than with the grid
(compare Time Cost, Figure\ref{RectangularGridTwoWayPlots} and \ref{RandGridTwoWayPlots}).

Overall, for 6D, 4x4, costs were substantially similar for both rectangular and random grids.

%%%%** Section 3.4.3 
\subsubsection{6D Normality}

{\it Rectangular Grid: } We analyzed the normality of the costs computed from 10,000 randomly sampled trajectories in the 6D 4x4 {rectangular} grid
(Figure \label{RectangularGridTwoWayPlots6D}) using the same Q-Q Plot methodology of Figure \ref{QQ4x41M}
indicating very Gaussian behavior of costs in the higher dimensional space (Figure \ref{6Dx4x4RectQQ}).

{\it Random Grid: } We also analyzed the normality of the costs computed from 10,000 randomly sampled trajectories in the 6D 4x4 {\it random} grid
(Figure \label{RandGridTwoWayPlots6D}) which was also highly Gaussian (Figure \ref{6Dx4x4RANDOMQQ}).

%%%%** Section 3.4.4 
\subsubsection{Ties}

Ties (instances where multiple branches from a node have about the
same cost) are an interesting feature that seems to be less frequently explored
in the literature.  Our NN searches of the phase space grids revealed that
tie situations are very common for both rectangular
and random grids (Figure \ref{RandPointTiesTimeEnergy}).  Prevalence of tied
nodes may suggest a large space for additional
heuristic improvements in the neighborhood of our paths discovered   by
the NN heuristic with random tie breaking.

As might be expected, the tie distribution curve is smoother for the random
grid and has more discrete features for the rectangular grid.
However for the time cost on the random grid, the smooth curve is abruptly truncated in the search
data at a maximum tie multiplicity of 76 (i.e. 76 branches with equal cost).
It may be
surprising that so many ties were found in the random grid.  We used a
2\% threshold for determining equal cost so this curve may be smaller if a
tighter tolerance were used.

%%%%** Section 4 
\section{Conclusions and Discussion}

This study has explored ways to most efficiently search a phase space of
up to 6 dimensions.  We compared nearest-neighbor (NN) searching  and random
sampling, first on a 2D phase space in which it was feasible to find the
global optimum for comparison and then on the 6D space with up to 4 points
per axis ($N=4$).

The motivation for this study is efficient collection of motion data over the
position and velocity workspace (phase space) of a physical robot for training
machine learning algorithms.
Efficient search of this space is important for proper coverage of robot
operating conditions, feasible training time, and avoidance of over-fitting.

Computation cost for optimal paths grew extraordinarily rapidly as the space
dimension approached practical applications.  However, when equal computation
time is applied to NN search and random sampling there is a vanishingly small
probability that the random samples contain a result as good as the NN search.
This is perhaps non-obvious since there is evidence \cite{Gutin2001TravelingSS} that NN algorithms
can produce bad results in some  TSP problems.

The distribution of costs from random sampling as well as from large numbers
of NN searches was very close to Gaussian.  This was expected from the
literature \cite{frieze2007probabilistic} and it allows estimates of the likelihood
that good results could be obtained by NN search results (compared to large random samples).
However a limitation of this analysis
is that, for the rectangular grids, the quantile-quantile plots from 1 million samples(Figure \ref{QQ4x41M})
have inflections in their negative extremes which may invalidate the use of $Z$
statistics.

In future work,
we intend to apply these algorithms to programming
actual robot motion for efficiently collecting mechanical accuracy training data with sufficient
ability to generalize throughout the phase space.

%%%%** Section 5 
\section{Appendix: Notation and Basics}\label{SecNotationBasics}

%%%%** Section 5.1
\subsection{Notation}
The goal is to search a grid of points in the  space consisting of points  $P_i = \{X_i, V_i\} = \{x,y,z,\dot{x},\dot{y},\dot{z}\}$ within bounds:
\[
-1 < \{x,y,z,\dot{x},\dot{y},\dot{z}\} < 1
\]

We wish to visit all the points with as low a cost as possible.
We will set up a grid with $N$ points per axis, for a total of $6^N$ points
or a set of $6^N$ random points .

A {\it trajectory}, $T_{ij}$ between two points in this space, $T(P_i,P_j)$, is a route through
the space from $P_i$ to $P_{i+1}$ with the properties

\beq \label{firstconstraint}
\Delta X (T_{ij}) = \frac{X_{i+1}-X_i}{\Delta t}
\eeq
%

%%%%** Figure 16 
\begin{figure}\centering
  \includegraphics[width=\columnwidth]{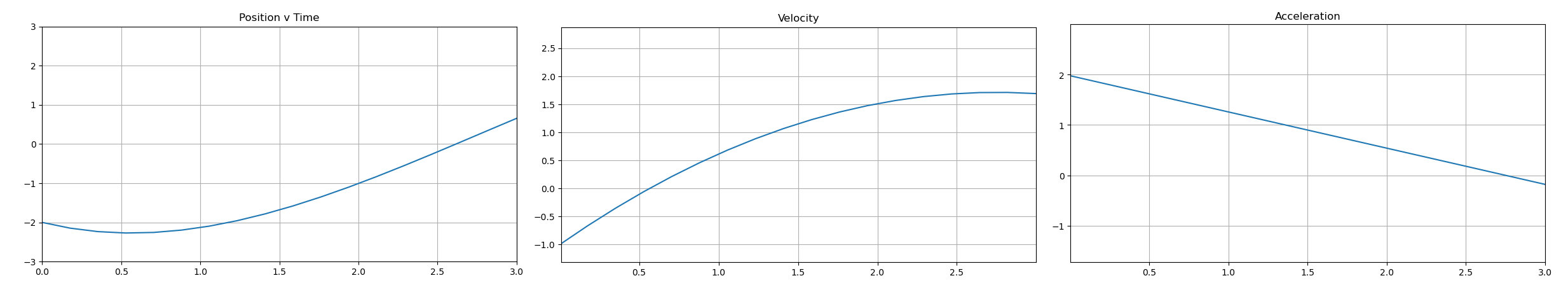}
  \caption{A trajectory between two example points in 2D phase space computed by the 3rd order polynomial.}\label{basicTraj}
\end{figure}

%%%%** Section 5.2
\subsection{Trajectories between points}\label{TrajectorySolved}

A {\bf Trajectory}, $T_{ij}$ connects point $P_i$ to point $P_j$
in phase space with a time function $x(t), 0<t<\Delta t $.  To meet the constraints
\beq \label{trajConstraints}
x(0) = X_i, \; x(\Delta t) = X_j, \; \dot{x}(0) = V_i, \;
\dot{x}(\Delta t) = V_j
\eeq
we can use a 3rd order polynomial having four unknown constants:
\beq
\begin{aligned} \label{polyTraj}
x(t) & = a_0 + a_1t + a_2t^2 + a_3t^3 \\
v(t) & = \hspace{7mm}  a_1+2a_2t + 3a_3t^2 \\
a(t) & = \hspace{16mm}  2a_2 + 6a_3 t \\
\end{aligned}
\eeq

A typical trajectory of this type, computed for
\beq
x(0) =  -2, \; v(0) = -1, \qquad  x(\Delta t) = 1.5, \; v(\Delta t) = 1.5
\eeq
is given in Figure \ref{basicTraj}.

The constants are solved as follows:
\beq
a_0 = x(0),  a_1 = v(0)
\eeq
defining some intermediate terms:

\beq
\Delta x = x(\Delta t)-x(0) \qquad \Delta v = v(\Delta t)-v(0)
\eeq

\beq
b0 = \Delta t \qquad b1 = {\Delta t}^2 \qquad b2 = {\Delta t}^3
\eeq

\beq
b3 = 2\Delta t \qquad b4 = 3{\Delta t}^2
\eeq
then
\beq
a_3 = \frac {b1 \Delta v - b3(\Delta x-v(0)b0)}  {b1b4-b2b3}
\eeq
\beq
a_2 = \frac  {\Delta x - v(0)b0 - a_3b2}   {b1}
\eeq
Our goal is to find a minimum cost trajectory satisfying eqn
(\ref{trajConstraints}) and with the form of eqn (\ref{polyTraj}).
Then we can define the cost of each trajectory between two phase-space points at least two ways:

%%%%** Section 5.2.1 
\subsubsection{Energy Cost}    We assume that energy of a trajectory is
\beq
C_{e}(T_{ij}) = \int_0^{\Delta t} a(t)^2 dt
\eeq
this can be solved using (\ref{polyTraj}) as
\beq
C_{e}(T_{ij}) = 4a_2(\Delta t) + 12a_2a_3(\Delta t)^2 + 12a_3^2(\Delta t)^3
\eeq

%%%%** Section 5.2.2 
\subsubsection{Duration Cost}   The time cost, $C_t$ is
\beq
C_t(T_{ij}) = \Delta t
\eeq

%%%%** Section 5.2.3 
\subsubsection{Acceleration Constraint}

To assure that our trajectories, $x(t),v(t),a(t)$ are feasible
for a real robot manipulator, we will constrain
\beq
|a(t)| < a_{max} \qquad 0<t<\Delta t
\eeq
Furthermore we wish to complete the trajectory as fast as
feasible, so we will set this constraint to equality:
\beq \label{acc_const}
\max(|(a(t)|) = a_{max}
\eeq
From eqn (\ref{polyTraj}) we know that acceleration is linear with
time for all solutions, thus we have:
\beq \label{acc_max}
\max(|a(t)|) = \max(|a(0)|, |a(\Delta t)| )
\eeq
We iteratively minimize $\Delta t$ for each trajectory until
eqn (\ref{acc_const}) is satisfied within about 2\%.

%%%%** Section 5.3
\subsection{Path Cost}
A {\it path}, $\mathbf{P}$, is a sequence of
trajectories (indexed by $k$), $T_{ijk}$,
connecting $P_i$ to $P_j$
such that the trajectories are connected, e.g.
\beq
P_j(T_{ijk}) = P_i(T_{ijk+1})
\eeq
the points $P_i$ covering the entire grid.
Let $C_k=C_x(T_{ijk})$ be the cost of the $k^{th}$ trajectory in
the path, $\mathbf{P}$.
The time   cost of visiting every point in the path is
\beq
C_T = \Sigma_k C_k  \qquad 0 \leq i < N^m
\eeq
For example, the total time cost of path ${P}_1$ would be
\beq
C_{Tp1} = \Sigma_k C_t(T_{ijk})
\eeq
where $T_{ijk}$ is the $k^{th}$ trajectory of path ${P}_1$.

%%%%** Section 6 
 \section{Appendix: Statistical Analysis of NN search vs. random sampling in 6D}\label{SecSampleStats}

In this section we evaluate the likelihood that random sampling, since it is
much faster computationally, could in fact produce a result as good as nearest-neighbor
heuristic searching if the same computational resources are applied.
In higher dimensions we don't know the global optimum path but NN searching
produces very low cost paths compared to random samples.
\cite{frieze2007probabilistic} performed a probabilistic analysis of the TSP,
but it was limited to symmetric and Euclidean problems.

We  use the Z-statistic in three ways to determine how much better the NN search result is
than choosing the lowest cost from a large number of random samples.  But first, we must
equalize the computing resources for a fair comparison.  Random sampling is much more efficient
than the NN search because  random paths are generated by permutations, and we only have to
evaluate the sum of all branch costs in the path ($O(n)$, where $n$ is the path length).
In contrast, with the NN search, at each node,
we have to evaluate the cost of all un-visited branches to get the minimum
cost value, and in our case search again to find the set of 'tied' nodes ($O(n^2)$).

We  experimentally determined that for
our 6D problem with grid size 3, 10 NN searches
took about the same computation time as did computing the costs of
10,000 random trajectories and selecting the lowest cost path.
We are therefore able to compare the two cases statistically.

The first Z-statistic is somewhat traditional.  If $\mu_{samp}$ and $\sigma_{samp}$
describe the 10,000 samples, and $\mu_{NN}$ is the mean of a set of NN results, we have
\beq
Z = \frac   {\mu_{NN}-\mu_{samp}}   {\sigma_{samp}}
\eeq
We could also define a similar statistic, $Z'$ in terms of the minimum value of the
two samples, since they are readily available computationally:
\beq
Z' = \frac   {min(NN)- min({sample})}   {\sigma_{samp}}
\eeq
and to evaluate the worst case,
\beq
Z'' = \frac   {max(NN)- min({sample})}   {\sigma_{samp}}
\eeq

We can view $\{Z, Z', Z''\}$ as multiples of $\sigma_{samp}$ to judge how ``rare" is
each NN search result.
These Z-scores are all negative for our data since NN searching gives lower cost than 10,000
samples in all experiments.

The probability, $P_c$, that a randomly selected path would have lower cost than
$Z$,  is
\beq\label{P_cEquation}
P_c(Z) = 1.0 - CDF(Z) = 1.0 - \int_{-\infty}^Z N(t) dt
\eeq
where $N(t)$ is the normal distribution.   We can make analogous definitions
for the min and max $Z$ scores, $Z',Z''$.

We consider the case where 10,000 random samples are drawn from a normal distribution.
For a given $Z$, what is the probability that a lower value will not turn up in the 10,000
samples?
The probability, $P_h$, that a single sample path will NOT have lower cost is
\beq
P_{h} = 1.0 - P_c(Z)
\eeq
If we draw $n$ times, and the samples are independent, then the probability
that a lower value will not be drawn, $P_{hn}$, is
\beq
P_{hn} = (P_h)^n
\eeq
Thus, the probability that $n$ random samples WILL include one with
a lower Z score is
\beq\label{P_ln}
P_{ln} = 1.0 - P_{hn} = 1.0 - (P_h)^n = 1.0 - (1.0-P_c(Z))^n
\eeq
Where $P_{ln}$ is the probability of drawing a lower cost sample in $n$ tries.

Using {\tt Python3}'s {\tt scipy.stats} package, the most negative integer Z score
that gives a non-zero cdf probability  is
$Z=-8$ which gives $P_c(-8)=6.661\times10^{-16}$.
Using this conservative Z value, and $n=10,000$, eqn (\ref{P_ln}) gives
$P_{hn} = 6.661\times10^{-12}$.

Thus the probability that 10,000 random searches would find a result as good
as the NN search is on the order of 1 in a trillion, and even lower for Z-scores below -8.

The results (Table \ref{10xSearchCompare}) show this comparison for the rectangular
grid (rows 1-4) and three different random grids (rows 5-16).
10 iterations (from different random starting points) are compared with
10,000 random path costs for both Time and Energy.  The table shows
\begin{itemize}
  \item $Z$ and $Z'$ statistic values for the NN searches range from $-45 \geq Z \geq -74$,
  dramatically  greater than the $Z=-8$ example above.
  \item $Z''$ statistics (not shown in table for reasons of space)
  were slightly lower than $Z$ and $Z'$ and ranged from -40 to -70.
  \item Computation times are approximately equivalent  (31-51sec).
  \item Results for 3 different random grids were nearly the same (rows 5-16).
\end{itemize}

We can thus conclude that the NN search is substantially more efficient than sampling.
There may be a feasible
computation time greater than 1 minute which might do as well as the NN search
(which does not seem to improve
much with increasing search sizes).  But the very large $Z$ scores suggest otherwise.

These results may be affected by the inflections of the Quantile-Quantile plot
(Figure \ref{QQ4x41M}) but since that plot includes all the data above, the
effect should be small and predominantly affect energy cost.

%%%%** Table 1 
\begin{table*}[!ht]
    \centering
    \begin{tabular}{|r|c|c|l|l|r|r|r|r|r|r|r|r|c|}
    \hline
        	{\bf Exp} & {\bf D} & {\bf N} & {\bf Grid} & {\bf Cost} & {\bf Srch} & {\bf N} & {\bf mu} & {\bf min} & {\bf sig} & {\bf Z} & {\bf Z'} & {\bf T(sec)} & {\bf hash} \\ \hline
        1 & 6 & 3 & rect & Time & samp & 10,000 & 2245 & 2186 & 16.5 & ~ & ~ & 51.0 & cbc78246 \\ \hline
        2 & 6 & 3 & rect & Energy & samp & 10,000 & 9412 & 8952 & 123.0 & ~ & ~ & 53.0 & 5470ef83 \\ \hline
        3 & 6 & 3 & rect & Time & NH & 10 & 1019 & 1013 & ~ & 74.3 & 71.1 & 40.0 & 3a9de6e6 \\ \hline
        4 & 6 & 3 & rect & Energy & NH & 10 & 3404 & 3372 & ~ & 48.8 & 45.4 & 38.0 & 4718555c \\ \hline
        5 & 6 & 3 & rand 73d954cc & Time & samp & 10,000 & 1784 & 1733 & 12.7 & ~ & ~ & 47.0 & cdb3c021 \\ \hline
        6 & 6 & 3 & rand 73d954cc & Energy & samp & 10,000 & 6885 & 6546 & 88.2 & ~ & ~ & 49.0 & 548ee970 \\ \hline
        7 & 6 & 3 & rand 73d954cc & Time & NN & 10 & 1024 & 1021 & ~ & 59.8 & 56.1 & 38.0 & dfcc9b7b \\ \hline
        8 & 6 & 3 & rand 73d954cc & Energy & NN & 10 & 1634 & 1612 & ~ & 59.5 & 55.9 & 31.0 & c3e8227b \\ \hline
        9 & 6 & 3 & rand 786d6950 & Time & samp & 10,000 & 1770 & 1718 & 12.6 & ~ & ~ & 47.0 & 1c1c207e \\ \hline
        10 & 6 & 3 & rand 786d6950 & Energy & samp & 10,000 & 6819 & 6504 & 84.0 & ~ & ~ & 47.5 & f954b7fd \\ \hline
        11 & 6 & 3 & rand 786d6950 & Time & NN & 10 & 1027 & 1023 & ~ & 59.2 & 55.3 & 39.2 & 54ed4a67 \\ \hline
        12 & 6 & 3 & rand 786d6950 & Energy & NN & 10 & 1754 & 1720 & ~ & 60.3 & 57.0 & 34.8 & a219c0bc \\ \hline
        13 & 6 & 4 & rand 6d5755b7 & Time & samp & 10,000 & 1799 & 1737 & 12.6 & ~ & ~ & 47.2 & d98ce58c \\ \hline
        14 & 6 & 4 & rand 6d5755b7 & Energy & samp & 10,000 & 6892 & 6545 & 86.4 & ~ & ~ & 46.0 & ded60da5 \\ \hline
        15 & 6 & 4 & rand 6d5755b7 & Time & NN & 10 & 1031 & 1028 & ~ & 61.0 & 56.3 & 40.0 & d48770a4 \\ \hline
        16 & 6 & 4 & rand 6d5755b7 & Energy & NN & 10 & 1806 & 1764 & ~ & 58.9 & 55.3 & 38.8 & e96bd1eb \\ \hline
    \end{tabular}
\caption{Computational experiments exploring practicality of NN search results.
For both Time and Energy costs, NN searching produces a dramatically better result than
the best randomly chosen path if computation time is approximately the same.
$Z''$ scores (not shown) were 45-70. Hash codes identify pertinent data files.}\label{10xSearchCompare}
\end{table*}

\newpage

%  Use name of bibliography files without .bib extension
\bibliography{cto}
\end{document}